\documentclass{article}

\usepackage{PRIMEarxiv}

\usepackage[numbers, sort, super]{natbib}
\usepackage{natmove}
\usepackage[utf8]{inputenc} 
\usepackage[T1]{fontenc}    
\usepackage[hidelinks]{hyperref} 
\usepackage{url}            
\usepackage{booktabs}       
\usepackage{amsfonts}       
\usepackage{nicefrac}       
\usepackage{microtype}      
\usepackage{fancyhdr}       
\usepackage{graphicx}       
\usepackage{algorithm}      
\usepackage[noend]{algpseudocode}  
\usepackage{amsmath}        
\usepackage{setspace}       
\graphicspath{{media/}}     
\usepackage{dsfont}         
\usepackage{subcaption}     
\usepackage[table]{xcolor}  
\usepackage{bibunits}
\usepackage{everypage, eso-pic}      

\makeatletter
\def\BState{\State\hskip-\ALG@thistlm}
\makeatother

\newcommand\FirstPageBackground{%
  \AddToShipoutPictureBG*{%
    \put(0,0){%
      \parbox[b][\paperheight]{\paperwidth}{%
        \vfill
        \centering
        \includegraphics[width=\paperwidth,height=\paperheight,%
                         keepaspectratio]{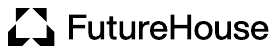}%
        \vfill
      }}}}

\newcommand\OtherPagesBackground{%
  \AddToShipoutPictureBG*{%
    \put(0,0){%
      \parbox[b][\paperheight]{\paperwidth}{%
        \vfill
        \centering
        \includegraphics[width=\paperwidth,height=\paperheight,%
                         keepaspectratio]{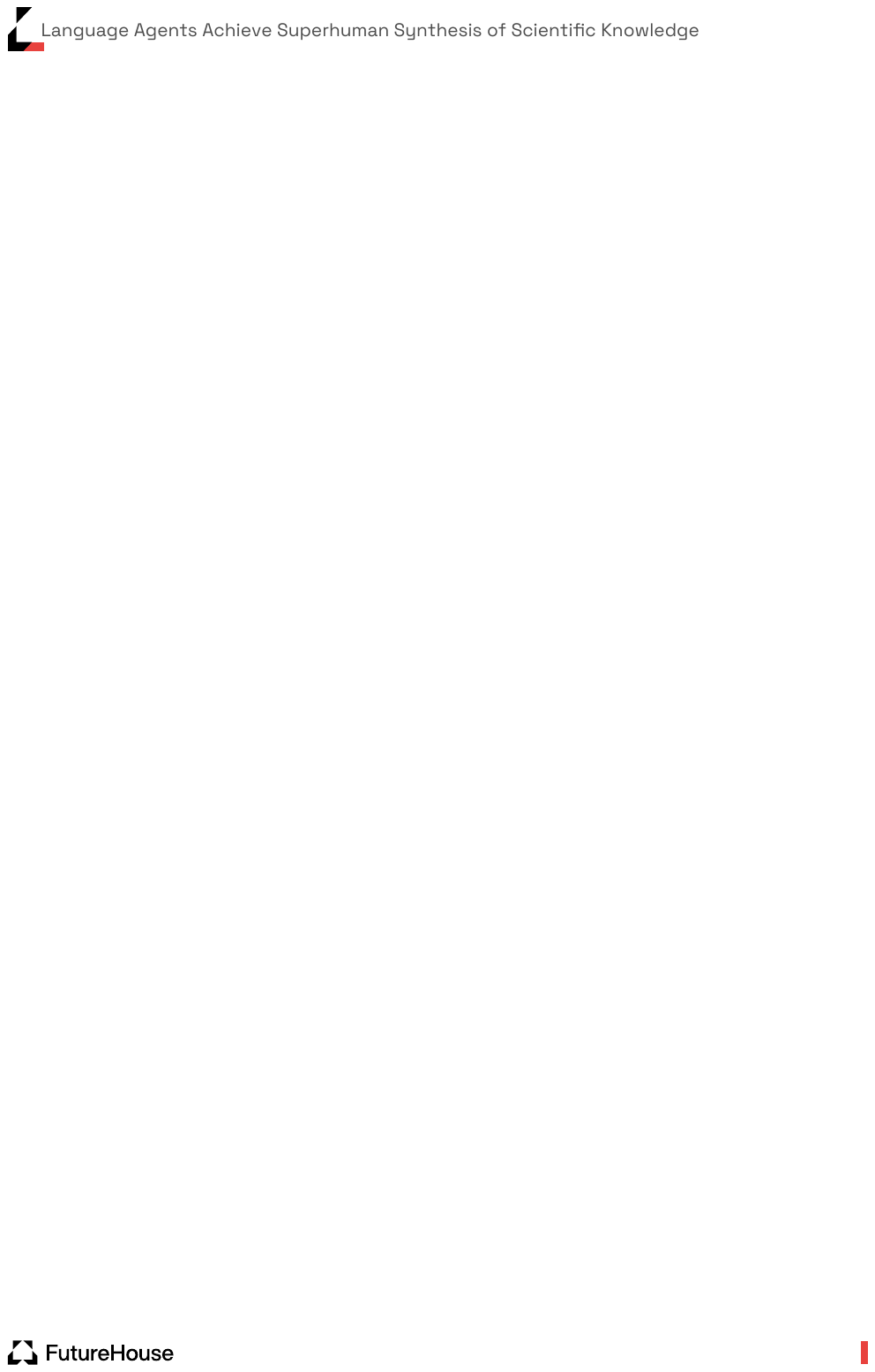}%
        \vfill
      }}}}

\AtBeginShipoutFirst{\FirstPageBackground}

\AddEverypageHook{%
  \ifnum\value{page}>0 %
    \OtherPagesBackground
  \fi}

\title{Language agents achieve superhuman synthesis of scientific knowledge
}

\author{
\begin{tabular}{ccc}
Michael D. Skarlinski$^{1}$ & Sam Cox$^{1,2}$ & Jon M. Laurent$^{1}$ \\ \\
James D. Braza$^{1}$ & Michaela Hinks$^{1}$ & Michael J. Hammerling$^{1}$\\ \\
Manvitha Ponnapati$^{1}$ & Samuel G. Rodriques$^{1,3*}$ & Andrew D. White$^{1,2*}$
\end{tabular}
\\
\\
$^{1}$FutureHouse Inc., San Francisco, CA\\
$^{2}$University of Rochester, Rochester, NY\\
$^{3}$ Francis Crick Institute, London, UK\\
$^{*}$These authors jointly supervise technical work at FutureHouse.\\
Correspondence to: \texttt{\{sam,andrew\}@futurehouse.org}
}

\begin{document}
\maketitle

\begin{abstract}
Language models are known to ``hallucinate'' incorrect information, and it is unclear if they are sufficiently accurate and reliable for use in scientific research. We developed a rigorous human-AI comparison methodology to evaluate language model agents on real-world literature search tasks covering information retrieval, summarization, and contradiction detection tasks. We show that PaperQA2, a frontier language model agent optimized for improved factuality, matches or exceeds subject matter expert performance on three realistic literature research tasks without any restrictions on humans (i.e., full access to internet, search tools, and time). PaperQA2 writes cited, Wikipedia-style summaries of scientific topics that are significantly more accurate than existing, human-written Wikipedia articles. We also introduce a hard benchmark for scientific literature research called LitQA2 that guided design of PaperQA2, leading to it exceeding human performance. Finally, we apply PaperQA2 to identify contradictions within the scientific literature, an important scientific task that is challenging for humans. PaperQA2 identifies $2.34 \pm 1.99$ (mean $\pm$ SD, $N=93$ papers) contradictions per paper in a random subset of biology papers, of which $70\%$ are validated by human experts. These results demonstrate that language model agents are now capable of exceeding domain experts across meaningful tasks on scientific literature. 
\end{abstract}
\vspace{0.7cm}



\begin{bibunit}[unsrt] 
\section{Introduction}
Large language models (LLMs) have the potential to assist scientists with retrieving, synthesizing, and summarizing the literature\cite{gao2023retrieval, shao2024assisting, Lo2023TheSR}, but still have several limitations for use in research tasks. Firstly, factuality is essential in scientific research, and LLMs hallucinate\cite{tonmoy2024comprehensive}, confidently stating information that is not grounded in any existing source or evidence. Secondly, science requires extreme attention to detail, and LLMs can overlook or misuse details when faced with challenging reasoning problems\cite{dahl2024large}. 
Finally, benchmarks for retrieval and reasoning across the scientific literature today are underdeveloped. They do not consider the entire literature, but instead are restricted to abstracts\cite{jin-etal-2019-pubmedqa}, retrieval on a fixed corpus\cite{krithara2023bioasq}, or simply provide the relevant paper directly\cite{jin2019pubmedqa}. These benchmarks are not suitable as performance proxies for real scientific research tasks, and, more importantly, often lack a direct comparison to human performance. Thus, it remains unclear whether language models and agents are suitable for use in scientific research.


\begin{figure}
    \centering 
    \includegraphics[width=0.8\textwidth]{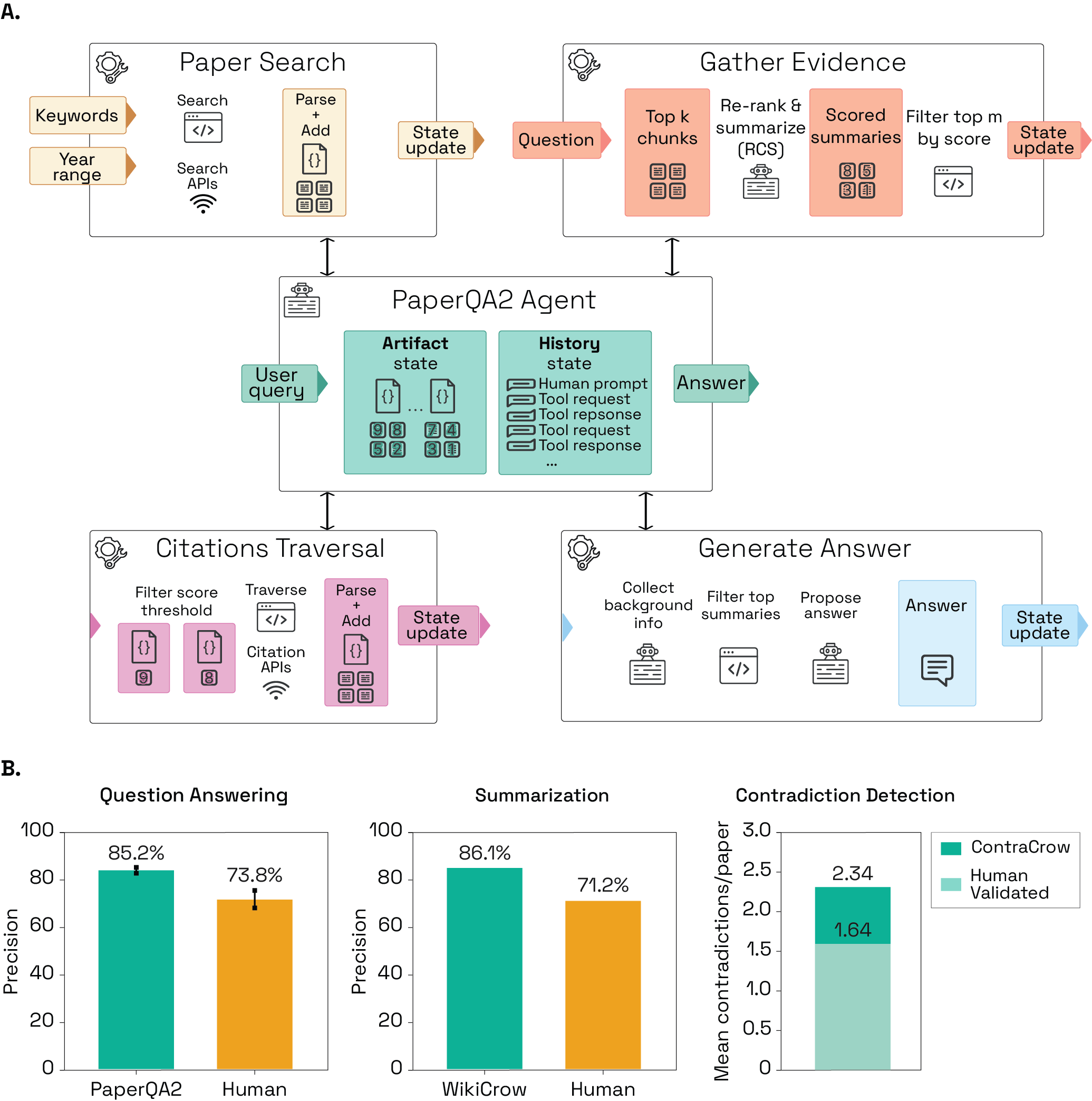}
    \caption{\textbf{A}. Schematic of PaperQA2's agentic toolset along with relevant action representations within each tool. \textbf{B.} PaperQA2 performance across question answering, cited article summarization, and contradiction detection. Error bars represent standard error.} 
    \label{fig:paperqa_summary} 
\end{figure}

We therefore set out to develop a rigorous comparison between the performance of AI systems and humans on three real-world tasks: a retrieval task involving searching the entire literature to answer questions; a summarization task involving producing a cited, Wikipedia-style articles on scientific topics; and a contradiction-detection task, involving extracting all claims from papers and checking them for contradictions against all of literature. This is, to our knowledge, the first robust procedure for evaluating a single AI system on multiple real-world literature search tasks. Using our newly developed evaluations, we explored multiple designs leading to a system we call PaperQA2 (\textbf{\autoref{fig:paperqa_summary}A}), which exceeds the performance of PhD students and postdocs on the retrieval and summarization tasks. Applying PaperQA2 to the contradiction detection task enables us to identify contradictions in biology papers at scale (\textbf{\autoref{fig:paperqa_summary}B}). For example, a statement that the ZNF804A rs1344706 allele positively affects brain structure in schizophrenia patients \cite{wang2019pleiotropic} was found to be contradicted by a later publication which found that rs1344706's effects on cortical thickness, surface area, and cortical volume in the brain aggravate the risk of schizophrenia\cite{wei2015znf804a}.

\section{Answering scientific questions}

\begin{figure}[h!]
    \centering
    \includegraphics[width=0.8\textwidth]{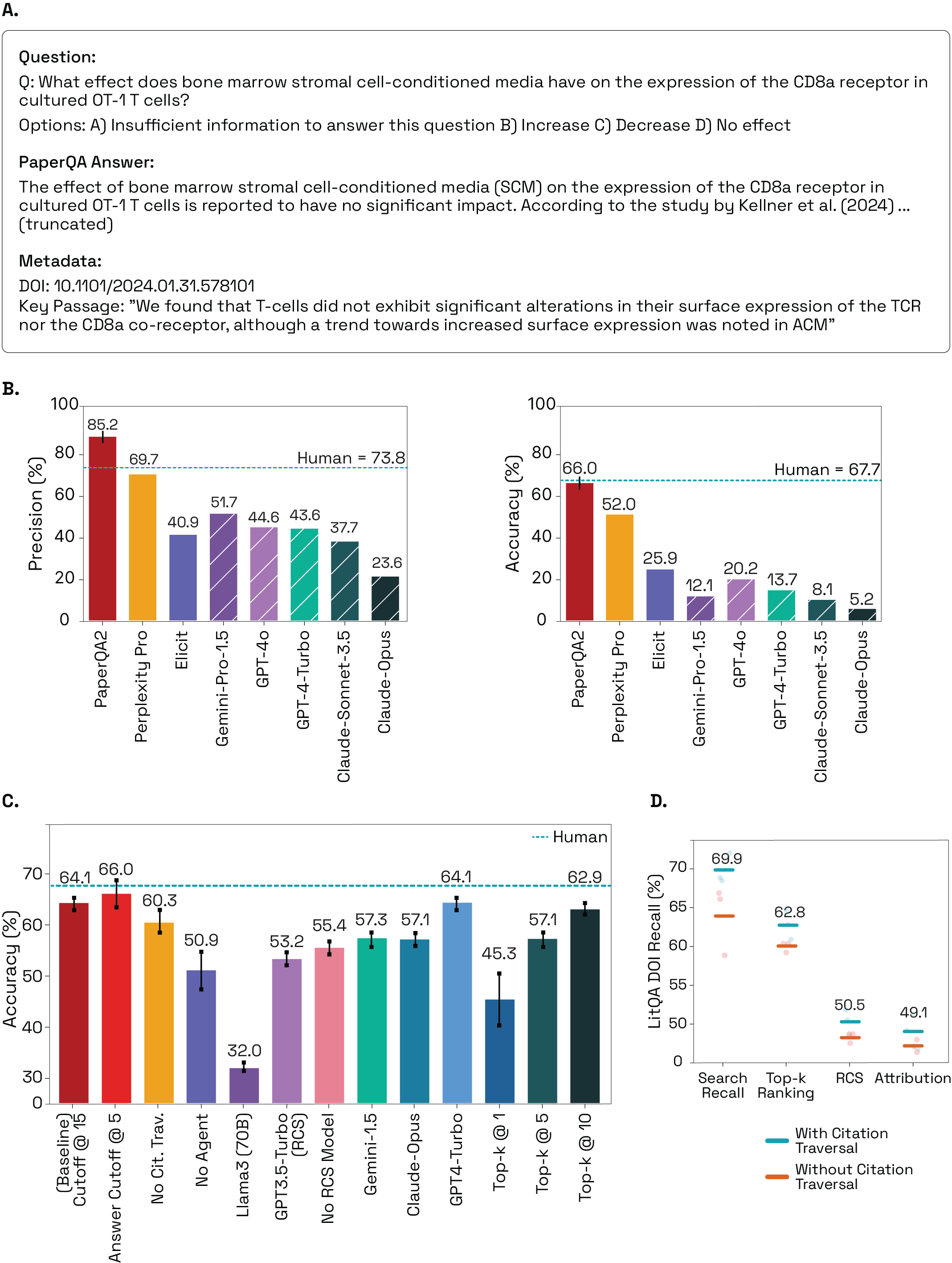}
    \caption{\textbf{A.} Example LitQA2 question, PaperQA2 answer, and metadata. \textbf{B.} PaperQA2 performance on LitQA2 across technologies. \textbf{C.} PaperQA2 performance studies and ablations across component categories. Error bars are 95\% CI. \textbf{D.} Aggregated LitQA2 DOI recall per PaperQA2 stage. \textit{Search Recall} includes DOIs found via the ``Paper Search'' or ``Citation traversal'' tools, \textit{Top-k Ranking} includes all DOIs with similarity rankings below the top-k ranking cutoff (30), \textit{RCS} includes all DOIs selected by RCS, and \textit{Attribution} includes all DOIs cited in the ``Generate Answer'' tool.}
    \label{fig:litqa_performance} 
\end{figure}

To evaluate AI systems on retrieval over the scientific literature, we first generated LitQA2,\cite{laurent2024lab} a set of 248 multiple choice questions with answers that require retrieval from scientific literature (\textbf{\autoref{fig:litqa_performance}A}). LitQA2 questions are designed to have answers that appear in the main body of a paper, but not in the abstract, and ideally appear only once in the set of all scientific literature. These constraints enable us to evaluate response accuracy by matching the system's cited source DOI with the DOI originally assigned by the question creator. To enforce these criteria, we generated large numbers of questions about obscure intermediate findings from very recent papers, and then excluded any questions where either an existing AI system or a human annotator could answer the question using an alternative source (\autoref{sec:question_evaluators}). These were generated entirely by experts, although there are emerging ideas about how to automate this process\cite{wan2024sciqag}. When answering LitQA2 questions, models can refuse to answer via selecting \textit{Insufficient information to answer this question}. Similar to prior work \cite{rajpurkar2018know} and matching actual scientific questions, some questions are intended to be unanswerable. We evaluate two metrics: precision, the fraction of questions answered correctly when a response is provided, and accuracy, the fraction of correct answers over all questions. We also consider recall, which is the total percentage of questions where the system attributed its answer to the correct source DOI denoted in LitQA2.

Having developed LitQA2, we then utilized it to design an AI system for the scientific literature. The current paradigm for eliciting factually-based responses from LLMs is to use retrieval-augmented generation (RAG)\cite{lewis2020retrieval, shuster-etal-2021-retrieval-augmentation}. RAG provides additional context to the LLM (e.g., snippets from research papers) to ground the generated response. As scientific literature is quite large, identifying the correct snippet is a challenge. Strategies like using metadata or hierarchical indexing can improve retrieval in this setting\cite{gao2023chatrec}, but finding the correct paper for a task often requires iterating and revising queries. Inspired by PaperQA\cite{lala2023paperqa}, PaperQA2 is a RAG \textit{agent} that treats retrieval and response generation as a multi-step agent task\cite{karpas2022mrkl} instead of a direct procedure. PaperQA2 decomposes RAG into \textit{tools}, allowing it to revise its search parameters and to generate and examine candidate answers before producing a final answer (\textbf{\autoref{fig:paperqa_summary}A}). PaperQA2 has access to a ``Paper Search'' tool, where the agent model transforms the user request into a keyword search that is used to identify candidate papers. The candidate papers are parsed into machine readable text, and chunked for later usage by the agent. PaperQA2 uses the state-of-the-art document parsing algorithm (Grobid\cite{GROBID}) that reliably parses sections, tables, and citations from papers. After finding candidates, PaperQA2 can use a ``Gather Evidence'' tool that first ranks paper chunks with a top-$k$ dense vector retrieval step, followed by an LLM reranking and contextual summarization (RCS) step. RCS prevents irrelevant chunks from appearing in the RAG context by summarizing and scoring the relevance of each chunk, which is known to be critical for RAG\cite{shi2023large}. The top ranked contextual summaries are stored in the agent's state for later steps. PaperQA2's design differs from similar RAG systems like Perplexity\cite{perplexity}, Elicit\cite{elicit}, or Mao et al.\cite{mao2020generation} which deliver retrieved chunks without substantial transformation in the context of the user query. While RCS is more costly than retrieval without a contextual summary, it allows PaperQA2 to examine much more text per user question. The RCS step also injects metadata about the source paper, like its citation count and journal. Once the PaperQA2 state has summaries, it can call a ``Generate Answer'' tool which uses the top ranked evidence summaries inside a prompt to an LLM for the final response to the asked questions or assigned task. To further improve recall, PaperQA2 adds a new ``Citation Traversal'' tool (\autoref{sec:cit-trav}) that exploits the citation graph as a form of hierarchical indexing to add additional relevant sources. 


In answering LitQA2 questions, PaperQA2 parsed and utilized an average of 14.5 $\pm$ 0.6 (mean $\pm$ SD, $n=3$) papers per question. Running PaperQA2 on LitQA2 yielded a precision of 85.2\% $\pm$ 1.1\% (mean $\pm$ SD, $n=3$), and an accuracy of 66.0\% $\pm$ 1.2\% (mean $\pm$ SD, $n=3$) (\textbf{\autoref{fig:litqa_performance}B}), with the system choosing ``insufficient information'' in 21.9\% $\pm$ 0.9\% (mean $\pm$ SD, $n=3$) of answers. To compare the performance of PaperQA2 to other retrieval systems, we evaluated the performance of PaperQA with original parameters, commercial systems like Perplexity\cite{perplexity} and Elicit\cite{elicit}, and frontier (non-RAG models) on LitQA2. We found that PaperQA2 outperforms other RAG systems on the LitQA2 benchmark in both precision and accuracy. We also found that all RAG systems tested, with the exception of Elicit, outperform non-RAG frontier models in both precision and accuracy. 

To ensure that we did not overfit PaperQA2 to achieve high performance on LitQA2, we generated a new set of 101 LitQA2 questions after making most of the engineering changes to PaperQA2. The accuracy of PaperQA2 on the original set of 147 questions did not differ significantly from its accuracy on the latter set of 101 questions, indicating that our optimizations in the first stage generalized well to new and unseen LitQA2 questions (\autoref{tab:litqa_before_after}).

To compare PaperQA2 performance to human performance on the same task, human annotators who either possessed a PhD in biology or a related science, or who were enrolled in a PhD program (see \autoref{sec:question_evaluators}), were each provided a subset of LitQA2 questions and a performance-related financial incentive of \$3-12 per question to answer as many questions correctly as possible within approximately one week, using any online tools and paper access provided by their institutions. Under these conditions, human annotators achieved 73.8\% $\pm$ 9.6\% (mean $\pm$ SD, $n=9$) precision on LitQA2 and 67.7\% $\pm$ 11.9\% (mean $\pm$ SD, $n=9$) accuracy (\textbf{\autoref{fig:litqa_performance}A}, green line). PaperQA2 thus achieved superhuman precision on this task ($t(8.6)=3.49, p=0.0036$) and did not differ significantly from humans in accuracy ($t(8.5)=-0.42, p=0.66$).

\section{Performance analysis of PaperQA2}
We varied the parameters of PaperQA2 to understand which are responsible for its accuracy (\textbf{\autoref{fig:litqa_performance}C}). We created a non-agentic version (\textit{No Agent}) which had a hard-coded sequence of actions (paper search, gather evidence, then generate answer). The non-agentic system had significantly lower accuracy ($t(3.7)=3.41, p=0.015$), validating the choice of using an agent. We attribute the performance difference to the agent's better recall because it can return to and change keyword searches (paper search tool calls) after observing the amount of relevant papers it finds. The highest accuracy LitQA2 runs had $1.26 \pm 0.07$ (mean $\pm$ SD) searches per question, and $0.46 \pm 0.02$ (mean $\pm$ SD) citation traversals per question showing that the agent will sometimes return to an additional search or traverse the citation graph to gather more papers.

\begin{figure}[h!]
    \centering 
    \refstepcounter{suppfigure}
    \includegraphics[width=0.7\textwidth]{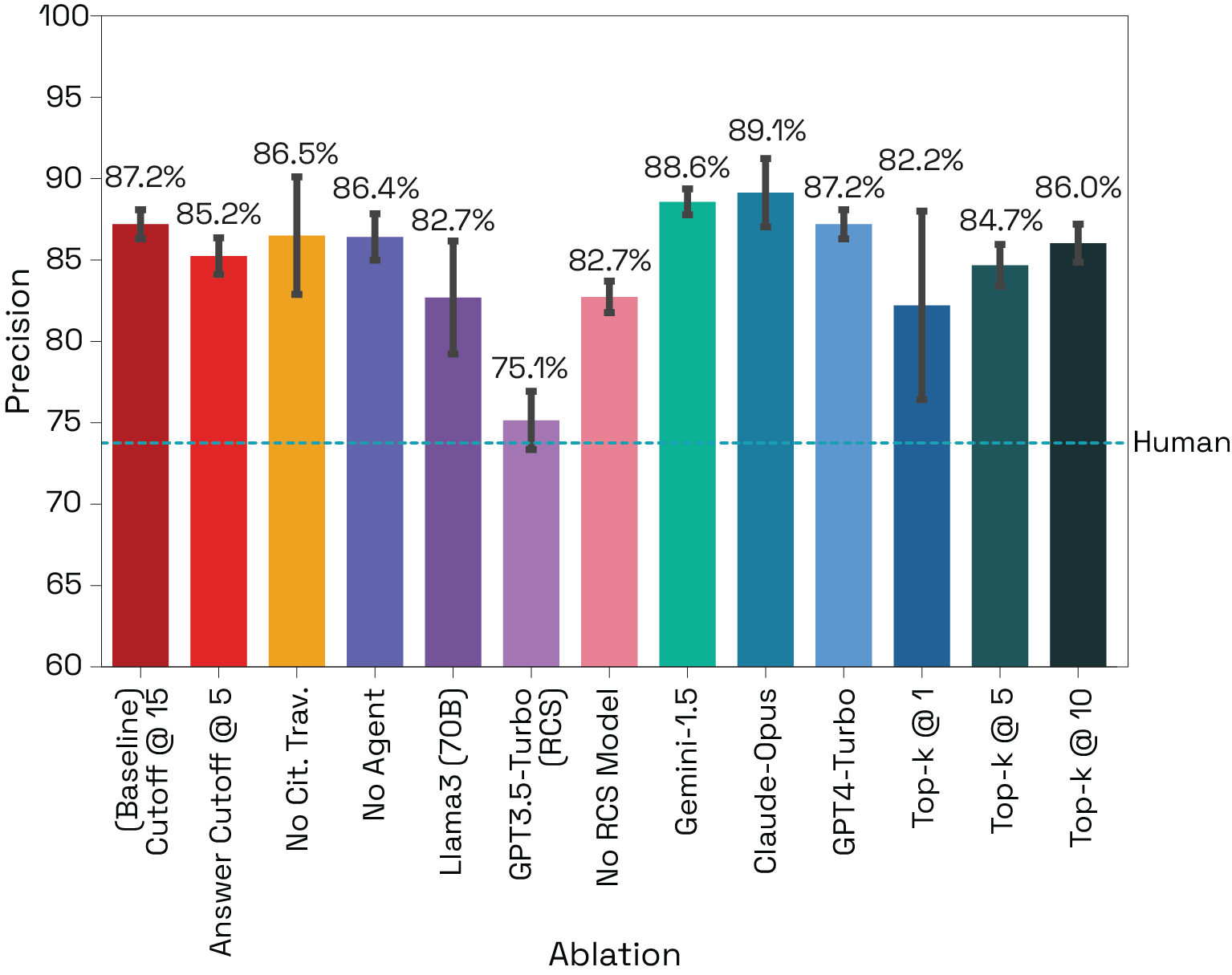}
    \caption{PaperQA2 precision performance on LitQA2 across all configuration categories included in \autoref{fig:litqa_performance}. All error bars are 95\% CI intervals.} 
    \label{fig:litqa_precision} 
\end{figure}

We then explored the tools and their parameters available to PaperQA2. An important RAG parameter is the number of contexts or text chunks to include in the final text generation step (``Generate Answer'' tool in \textbf{\autoref{fig:paperqa_summary}A}). This is a balance because increasing the count improves the chance of including the key context needed to answer a question (improving accuracy), but also increases the amount of distracting irrelevant context that reduces precision\cite{liu2023lostmiddlelanguagemodels}. We varied the amount of contexts from 15 to 5 in \textbf{Figure S\ref{fig:litqa_precision}} and \textbf{\autoref{fig:paperqa_summary}} to see this effect; 15 gives highest precision and 5 gives highest accuracy. 

To improve relevant chunk retrieval, we hypothesized that papers found as either citers or citees of existing relevant chunks would be an effective form of hierarchical indexing. This was validated by ablating the ``Citation Traversal'' tool (\textit{No Cit. Trav.}), which showed an increased accuracy ($t(2.55)=2.14, p=0.069$), and significantly increased DOI recall ($t(3)=3.4$, $p=0.022$) at all stages of the PaperQA2 flow. (\textbf{\autoref{fig:paperqa_summary}D}) This tool's process mirrors the way that scientists interact with the literature.

To quantify retrieval accuracy changes across LLM implementations, we performed experiments which varied the model choice for our ``Generate Answer'' tool as well as the RCS step in our ``Gather Evidence'' tool. Both of these tools give LLMs the opportunity to correctly identify crucial information in our scientific corpus, and we wanted to evaluate if their combined usage would be more effective than an un-transformed insertion of the top chunks into the final context window. The \textit{No RCS Model} ablation validates that adding RCS to a traditional RAG text generation step significantly increases retrieval accuracy ($t(3.92)=9.29, p < 0.001$). Interestingly, this is not true across all models tested, smaller models (GPT-3.5-Turbo (RCS), Llama3 (70B)) decrease overall accuracy when used for RCS, relative to not using a model at all. This indicated there is a comprehension threshold that must be met for effective summarization and relevance evaluation. GPT-4-Turbo significantly outperformed other models on LitQA2 accuracy when used in the RCS step ($t(3.47)=6.14, p=0.003$). Finally, Claude-Opus \cite{claude2024} had the highest LitQA2 precision (\textbf{\autoref{fig:litqa_precision}}), though it was not a significant increase over Gemini-1.5-Pro and GPT-4-Turbo.

To reduce the large cost of generating contextual summaries, we examined the effect of lowering the top-k ranking depth, i.e. limiting the number of passages considered in the RCS step. We see a significant increase in accuracy with increasing depth (i.e. more document chunks entering into the RCS ranking) from 1 to 10 ($t(2.15)=5.44, p=0.014$), and a long tail of diminishing performance gains from 10 to 30 (the default). In summary, having a deep (>10) RCS ranking list, with a high-performing LLM, was crucial in achieving human-level accuracy on LitQA2. 

We had hypothesized that parsing quality would affect accuracy, but Grobid parsings and larger chunk sizes did not significantly increase precision, accuracy, or recall on LitQA2 (\autoref{fig:chunksize_ablations}). This is likely specific to being a retrieval task, as there is often only a single passage needed from a paper's body, which makes our result insensitive to parser changes. Anecdotally, we found better parsings to be crucial for extracting data from tables in WikiCrow (detailed in the \autoref{sec:si-wikicrow}).

\section{Summarizing scientific topics} \label{sec:wikicrow}

\begin{figure}[h!]
    \centering 
    \includegraphics[width=0.8\textwidth]{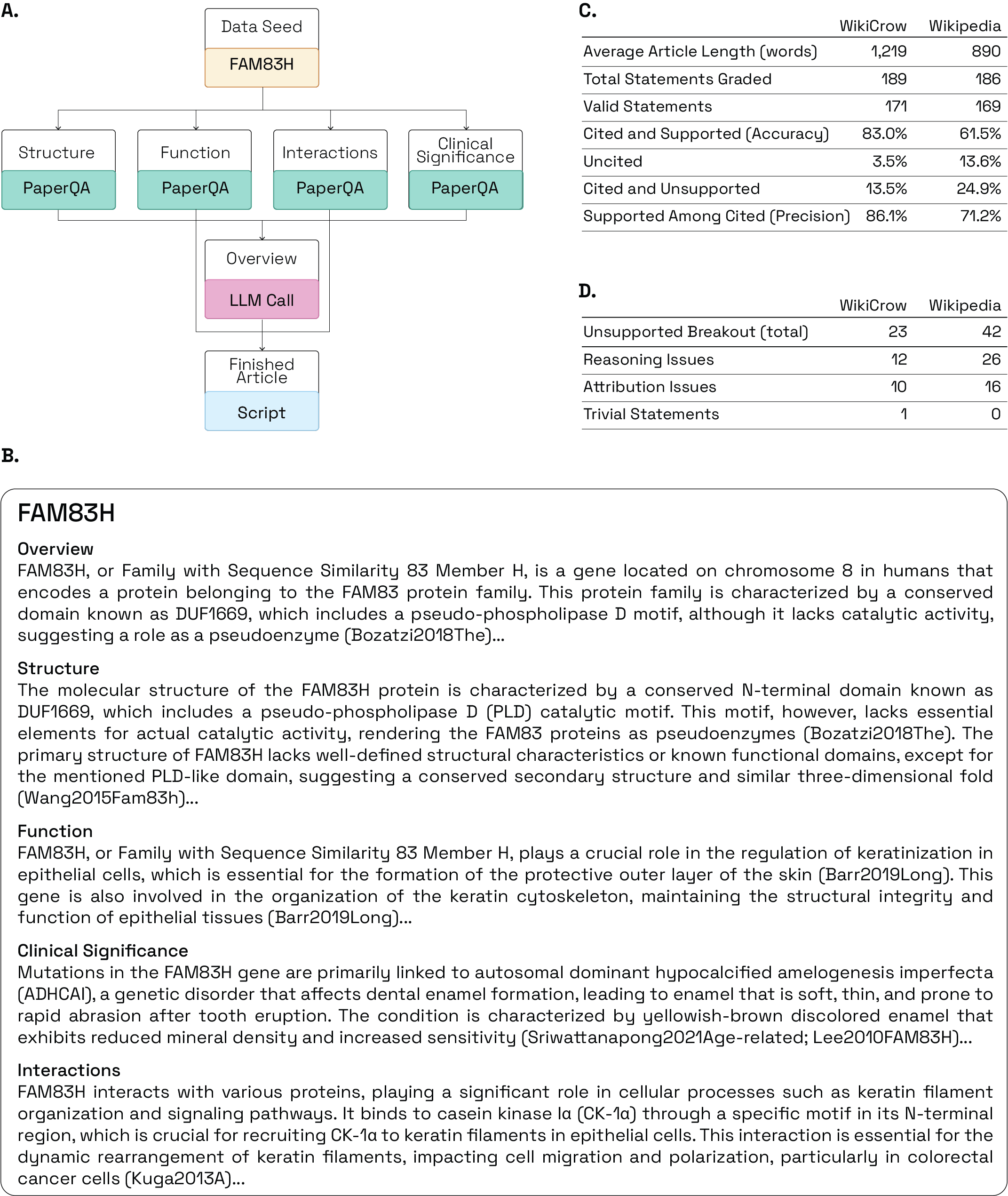}
    \caption{\textbf{A.} WikiCrow prompt graph, showing the initial seed variable (gene name), PaperQA2 prompts for each section, and overview LLM call. Each section feeds into a Python script which stitches the outputs together. \textbf{B.} Sample WikiCrow article for gene FAM83H, with truncated sections under each header. \textbf{C.} Performance statistics for a WikiCrow vs. Wikipedia comparison performed by evaluators. \textbf{D.} ``Cited and unsupported'' issue categorization counts.}
    \label{fig:WikiCrow_results} 
\end{figure}

To evaluate PaperQA2 on summarization, we engineered a system called WikiCrow, which generates cited Wikipedia-style articles about human protein-coding genes by combining several PaperQA2 calls on topics such as the structure, function, interactions, and clinical significance of the gene (\textbf{\autoref{fig:WikiCrow_results}A}). There has been previous work on \textit{unconstrained} document summarization, where the document must be found and then summarized,\cite{Giorgi2022ExploringTC} and even writing Wikipedia-style articles with RAG\cite{shao2024assisting}. These studies have not compared directly against Wikipedia with human evaluation. Instead, they used either LLMs to judge or compared ROGUE (text overlap) against ground-truth summaries. Here, we measure directly against human-generated Wikipedia with subject mater expert grading.

We used WikiCrow to generate 240 articles on genes that already have non-stub Wikipedia articles to have matched comparisons. WikiCrow articles averaged 1219.0 $\pm$ 275.0 words (mean $\pm$ SD, $N=240$), longer than the corresponding Wikipedia articles (889.6 $\pm$ 715.3 words). The average article was generated in 491.5 $\pm$ 324.0 seconds, and had an average cost of \$4.48 $\pm$ \$1.02 per article (including costs for search and LLM APIs). We compared WikiCrow and Wikipedia on 375 statements sampled from the 240 paired articles. Statements were selected using cues from document formatting (\autoref{sec:si-wikicrow}). The initial article sampling excluded any Wikipedia articles that were ``stubs'' or incomplete articles. Statements were then shuffled and given, blinded, to human experts, who graded statements according to whether they were (1) cited and supported; (2) missing a citation; or (3) cited and unsupported. We found that WikiCrow had significantly fewer ``cited and unsupported'' statements than the paired Wikipedia articles (13.5\% vs. 24.9\%) ($p=0.0075, \chi^2(1), N=375$ for all tests in this section). WikiCrow failed to cite sources at a 3.9x lower rate than human written articles, as only 3.5\% of WikiCrow statements were uncited, vs. 13.6\% for Wikipedia ($p < 0.001$). In addition, defining precision for WikiCrow as the ratio of cited and supported statements over all cited statements, we found that WikiCrow displayed significantly higher precision than human-written articles (86.1\% vs. 71.2\%, $p = 0.0013$).

The ``cited and unsupported'' evaluation category includes both inaccurate statements (e.g. true hallucinations or reasoning errors) and statements that are accurate with inappropriate citations. To investigate the nature of the errors in Wikipedia and WikiCrow further, we manually inspected all reported errors and attempted to classify the issues as follows: reasoning issues, i.e. the written information contradicts, over-extrapolates, or is unsupported by any included citations; attribution issues, i.e. the information is likely supported by another included source, but either the statement does not include the correct citation locally or the source is too broad (e.g. a database portal link); or trivial statements, which are true passages, but overly pedantic or unnecessary (\textbf{\autoref{fig:WikiCrow_results}D}). Surprisingly, we found that compared to Wikipedia, WikiCrow had significantly fewer reasoning errors (12 vs. 26, $p = 0.0144, \chi^2(1), N=375$) but a similar number of attribution errors (10 vs. 16, $p = 0.21$), suggesting that the improved factuality of WikiCrow over Wikipedia was largely due to improvements in reasoning. Although language models are clearly prone to reasoning errors (or hallucinations), in our task at least they appear to be less prone to such errors than Wikipedia authors or editors. This statement is specific to the agentic RAG setting presented here: language models like GPT-4 on their own, if asked to generate Wikipedia articles, would still be expected to hallucinate at high rates.
\footnote{All WikiCrow generated articles for this study are now available for download from a Google Cloud Storage bucket \texttt{https://storage.googleapis.com/fh-public/wikicrow2/}. A command line tool like \texttt{gsutil} can be used to list and bulk-access these files. All WikiCrow and LitQA evaluator responses are available via the following \href{https://docs.google.com/spreadsheets/d/1QvfxHSplVb-YyyDBkko_mhmbcomfOjBNtuW06f87jd0/edit?usp=sharing}{link}.}

\section{Detecting contradictions in the literature}

\begin{figure}[h!]
    \centering
    \includegraphics[width=0.8\textwidth]{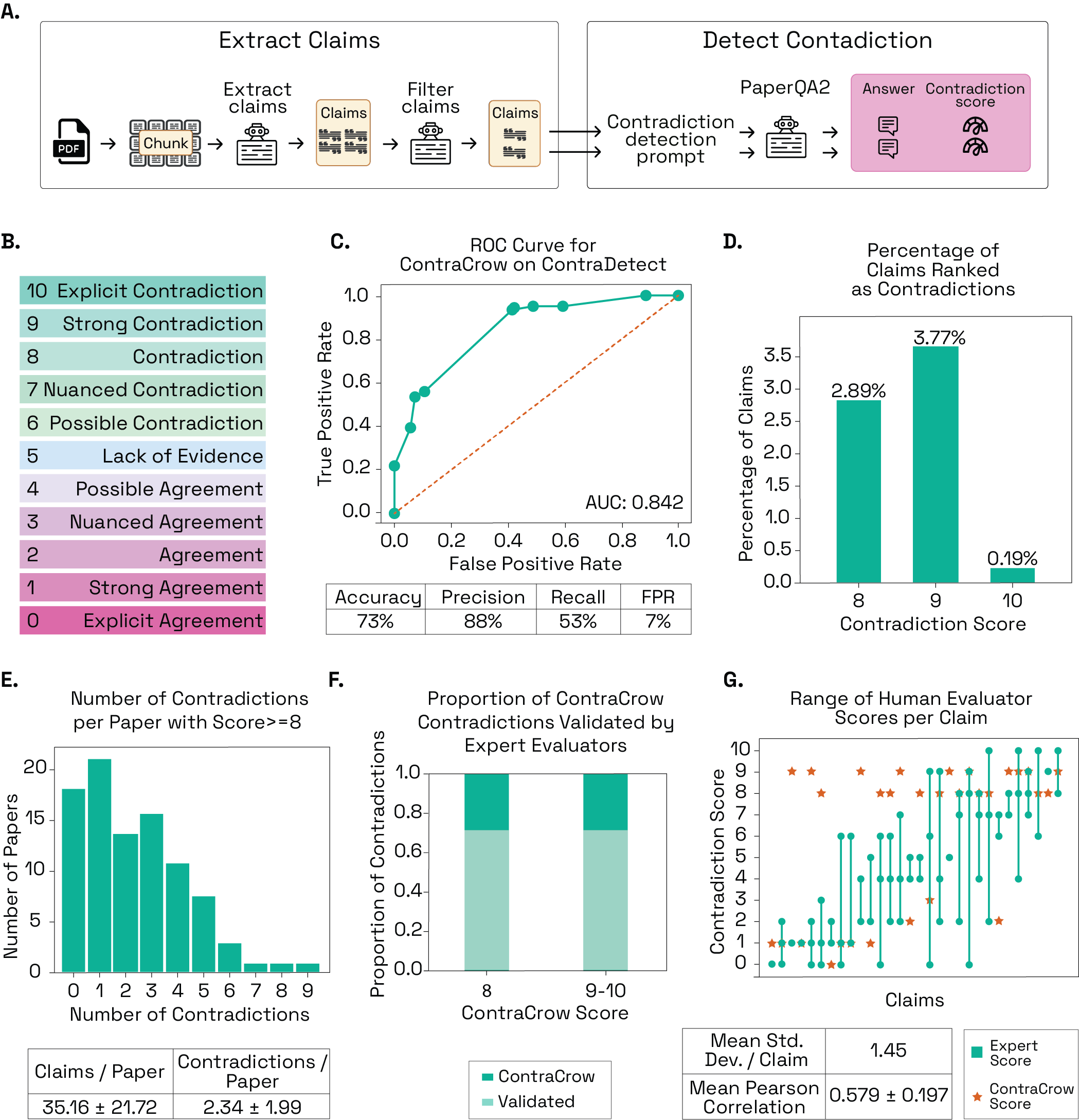}
    \caption{\textbf{A.} Schematic of ContraCrow. \textbf{B.} Likert scale used to evaluate contradictions, along with the integer mapping. \textbf{C.} ROC curve and metrics (Likert scale threshold of 8) of ContraCrow's performance on ContraDetect benchmark. \textbf{D.} Bar plot showing percentage of all claims found in 93 papers that are given scores of 8, 9, and 10 by ContraCrow. \textbf{E.} Histogram of the number of contradictions found per paper over 93 papers. The table shows the average number of claims ($35.16 \pm 21.72$ [mean $\pm$ SD, $N=93$]) and contradictions ($2.34 \pm 1.99$ [mean $\pm$ SD, $N=93$]) per paper found. \textbf{F.} Proportion of ContraCrow contradictions with scores 8 and 9-10 validated by expert evaluators. \textbf{G.} Range of evaluator scores for each claim. Bars represent the range of Likert scale scores assigned by expert evaluators and stars represent the ContraCrow label on the same claims. The table shows a mean standard deviation per claim of 1.45 ($N=30$ claims) over all evaluators and a mean Pearson correlation coefficient of $0.579 \pm 0.197$ (mean $\pm$ SD) between evaluators ($N=5$).}
    \label{fig:contradiction_detection}
\end{figure}

Because PaperQA2 can explore scientific literature at much higher throughput than human scientists, we reasoned that we could deploy it to systematically identify contradictions and inconsistencies in the literature at scale. Contradiction detection is a ``one versus many'' problem, which in principle involves comparing claims or statements in one paper with all other claims or statements in the literature. At scale, contradiction detection becomes a ``many versus many'' problem and loses feasibility for humans. Thus, we leveraged PaperQA2 to build a system called ContraCrow that automatically detects contradictions in the literature (\textbf{\autoref{fig:contradiction_detection}A}). 

Contradiction detection is also known as claim verification or colloquially as ``fact checking''\cite{vlachos2014fact}. This task has been studied for over a decade, especially in the context of claims in the news or the internet\cite{vlachos2014fact, ferreira2016emergent}. Although originally restricted to context and a claim, the setting extended to be unconstrained\cite{popat2017truth} and recent work tries to work at the scale of the internet\cite{schlichtkrull2024averitec}. Some claim verification work has also focused on scientific claims\cite{wadden2020fact, wadden2021multivers}. The main novelty of this work is in detecting contradictions without a restricted corpus and evaluating with human experts, not against a benchmark. 

ContraCrow first extracts claims from a provided paper using a series of LLM completion calls (similar to Schlichtkrull et al. (2024)\cite{schlichtkrull2024averitec}), and then feeds those claims into PaperQA2 with a \textit{contradiction detection} prompt. This prompt instructs the system to evaluate whether there are any contradictions in literature to the provided claim, providing both an answer and a choice from an 11-point Likert scale (\textbf{\autoref{fig:contradiction_detection}B}, Methods \autoref{sec:cc-methods}). Utilizing a Likert Scale allows the system to give more reliable and interpretable scores when providing rank\cite{abeysinghe2024challenges}.

To evaluate ContraCrow, we first derived a contradiction detection benchmark, ContraDetect, from LitQA2, as detailed in \autoref{sec:cc-eval}. Briefly, we converted half of the question-answer pairs in LitQA2 into declarative, incorrect statements that are contradicted by the papers referenced in the corresponding LitQA2 question. (For example, a question like ``what color is grass?'' would become ``grass is purple.'') We converted the other half of the LitQA2 questions into declarative, correct statements that are supported by the corresponding papers (``what color is grass'' becomes ``grass is green.''). 

We then evaluated ContraCrow on its ability to detect the contradictions in ContraDetect. By transforming the Likert scale output into integers, we were able to tune the detection threshold and obtain an ROC curve with an AUC of $0.842$ (\textbf{\autoref{fig:contradiction_detection}C}). Setting a threshold of 8 (\textit{contradiction}), ContraCrow achieved $73\%$ accuracy, $88\%$ precision, and a false positive rate of only $7\%$. To further evaluate precision and to demonstrate ContraCrow's ability to handle ``Has anybody ever done x?'' questions, we then ran ContraCrow on 42 ``no-evidence statements'' that have never to our knowledge been reported on (detailed in \autoref{sec:cc-eval}). These claims were hand-generated by the authors within their fields of expertise, and ContraCrow correctly chose \textit{lack of evidence} (5) as its response $98\%$ of the time, indicating its ability to distinguish between real contradictions and lack of support.

We then applied ContraCrow to a set of 93 biology-related papers randomly selected from our database, identifying an average of $35.16 \pm 21.72$ (mean $\pm$ SD, $N=93$) claims per paper. Of the 3,180 claims analyzed over the 93 papers, $6.85\%$ were deemed by ContraCrow to be contradicted by the literature, with $2.89\%$, $3.77\%,$ and $0.19\%$ assigned scores $8$, $9$, and $10$, respectively (\textbf{Figure \ref{fig:contradiction_detection}D}). Setting a Likert scale threshold of $8$, we detected an average of $2.34 \pm 1.99$ contradictions per paper (mean $\pm$ SD) (\textbf{Figure \ref{fig:contradiction_detection}E}). As an example, one contradiction detected by our system concerned the prognostic implications of LEF1 expression in colorectal carcinomas. The source paper, Kriegl et al. (2010)\cite{kriegl2010lef}, finds using immunohistochemistry on a tissue microarray that LEF-1 correlates positively with longer overall survival. By contrast, a study published the following year also using tissue microarrays found LEF1 overexpression in colorectal cancer correlates negatively with longer overall survival and is also correlated with liver metastasis\cite{lin2011comparative}, results that have been supported by other studies as well\cite{wang2013increased,wang2013knockdown,kermanshahi2014lef}, thus explicitly contradicting the original statement.

To evaluate the validity of the contradictions detected this way, expert human annotators evaluated 50 claims that were assigned ContraCrow scores of 8 and 50 claims that were assigned scores of 9 or 10, considering all evidence and reasoning cited by the model. Annotators agreed with ContraCrow's findings on $70\%$ of evaluated claims, or 1.64 contradictions per paper, demonstrating significant agreement ($p = 1e^{-4}, \chi^2(1)$), with an F1 score of $0.82$. Interestingly, claims scored as 8 by ContraCrow were not more likely to elicit human agreement (70\%) than claims scored as 9 or 10 (70\%) (\textbf{Figure \ref{fig:contradiction_detection}F}). The final number, $1.64$ validated contradictions per paper, serves as a lower bound on the abundance of human-validatable contradictions in the biology literature.

We were concerned that the annotators on the ``contradiction validation'' task might exhibit some bias toward agreeing with the model or otherwise be influenced by the model's reasoning and chosen sources. We therefore further evaluated ContraCrow by asking human experts to identify contradictions in the same literature considered by ContraCrow, without any access to ContraCrow's reasoning. This ``contradiction detection'' task (\autoref{sec:cc-eval}) is a much more challenging task for annotators than the contradiction validation task, because it requires them to exhaustively consider a significant amount of literature relevant to the claim in question, whereas the contradiction validation task only requires them to consider the specific sections cited by the model. In the contradiction detection task, human experts were provided with the top 15 paper chunks identified as relevant to the claim by PaperQA2 (the same chunks ContraCrow had access to), and were asked to respond whether the claim was contradicted by the available evidence on both the Likert scale and binary determination (``yes'' or ``no''). Experts considered a mixture of claims rated as contradictions and non-contradictions by ContraCrow. We found that experts' binary responses agree with each other on $75.5\% \pm 13.43\%$ of claims provided (mean $\pm$ SD, $N=10$ pairwise comparisons between human annotators), whereas they agree with ContraCrow on $60.42\% \pm 5.99 \%$ of claims (mean $\pm$ SD, $N=5$ human annotators), indicating that humans are significantly more correlated with each other than they are with ContraCrow ($p=0.015$). Examining the Likert scale values over 30 claims evaluated by human annotators leads us to speculate that overconfidence on ContraCrow's part is the primary driver of its lack of agreement with human annotators, and indicates directions for future improvement (\textbf{Figure \ref{fig:contradiction_detection}G}).

We are thus able to establish a lower bound for the average number of human-validatable contradictions per paper in biology literature at 1.64. Importantly, just because a claim is detected as a contradiction does not mean that the claim is wrong. For example, one detected contradiction is \textit{``GBP is only found in the cytosol of human fibroblasts.''}\cite{cheng1985affinity} This example shows how scientific literature and findings can update and contradict over time, as ContraCrow points out: \textit{``...more recent research has shown that GBPs can localize to various cellular compartments and membranes\cite{britzen2010intracellular, modiano2005golgi, naschberger2006human, schelle2023functional}.''} Thus, claims can contradict but still be contextually valid, and contradictory claims may be merely reflective of the iterative nature of research.

\section{Conclusions}

We developed a methodology to compare or validate AI systems against human performance in realistic tasks for scientific research. PaperQA2 outperforms human experts on answering questions across all scientific literature; produces summaries that are, on average, more factual than Wikipedia summaries; and can be deployed to identify contradictions in scientific literature at scale. The contradiction work, in particular, attests to the potential of systems such as PaperQA2 for science: notably, one human expert who performed this task, who was also tested on a large battery of other benchmarks, reported without solicitation that the contradiction detection task was the hardest task they were asked to perform. Although PaperQA2 is expensive compared to lower accuracy commercial systems, it is inexpensive in absolute terms, costing \$1 to \$3 per query. Scaling up PaperQA2 and other literature-enabled agents like WikiCrow and ContraCrow empowers us to take advantage of the latent knowledge in literature at much greater scale than is possible today.

\section{Data Availability}

The code necessary to replicate these results or modify the algorithm for further research is included on Github via \href{https://github.com/Future-House/paper-qa}{paperqa}. Data including all evaluator responses, contradiction detection claims, litQA questions, and WikiCrow candidate statements are available in the supplementary materials. All generated WikiCrow articles for this study are available in a public Google Cloud bucket here: \texttt{https://storage.googleapis.com/fh-public/wikicrow2/}.

\paragraph{Author Contributions} MDS led work on PaperQA2 and WikiCrow, in collaboration with JDB. SC led work on ContraCrow. JML led work on LitQA. MH, MJH, and MP contributed questions to the LitQA benchmark. SGR and ADW conceived of and supervised the project.

\paragraph{Acknowledgments} Work at FutureHouse is supported by the generosity of Eric and Wendy Schmidt. We also acknowledge Matt Rubashkin for contributions to an earlier stage of the project, and we acknowledge all current members of FutureHouse for useful discussions, including Sid Narayanan, Ryan Rhys-Griffiths, Cade Gordon, Peter Chang, and Conor Igoe. We also acknowledge support and resources from the Semantic Scholar Project at the Allen Institute for AI.

\paragraph{Competing interests} The authors report no competing interests. FutureHouse, Inc. is a non-profit research organization. 

\putbib[references] 

\end{bibunit}

\newcounter{suppfigure}
\renewcommand{\thesuppfigure}{S\arabic{suppfigure}}
\setcounter{suppfigure}{0}

\section{Methods}

\subsection{PaperQA Implementation and Parameters}

All reported figures and data in this work were built on the open source PaperQA package, available on GitHub at \href{https://github.com/Future-House/paper-qa}{paperqa}. While the core PaperQA repository provides the basic algorithms used, it does not include the Grobid parsing code, access to non-local full-text literature searches, or the citation traversal tool. The open source version of PaperQA utilizes LangChain \cite{LANGCHAIN} for its agentic and state update operations. The full configuration objects for all experiments run in this paper are included for further customization.

Note that while \texttt{paperqa} gives the ability to recreate this work, the experiments reported in this paper were performed using a more featureful HTTP server that takes advantage of bespoke infrastructure at the authors' institution. This infrastructure includes features such as user authentication, MongoDB request caching, Redis object caching and global-load balancing, several PostgreSQL DBs with associated ORM code, cost-monitoring modules, time-profiling modules, configuration storage and run orchestration (Dagster\cite{dagster} and kubernetes\cite{kubernetes}), cloud bucket storage for PDFs, a CI pipeline with semi-automated deployments, and infrastructure code for deploying auto-scaling instances in the cloud. None of these features affect performance on a per-query basis, but provide increased scalability, measurability, and persistence. To run our same server infrastructure, users would need to provision all of these assets and configure the deployments themselves. \texttt{paperqa} should serve allow usage and customization sufficient for most research purposes, and should be sufficient to reproduce the results reported here.

Even within \texttt{paperqa}, the ``Paper Search'' tool is limited by access to full text repositories of scientific papers, often bound by licensing agreements. The included implementation only works from local files accessible to each user. Our implementation starts with a full or partial-text keyword search, where the keywords have been specified by the agent when selecting the paper search tool. The ranked results returned from these services are then matched to a user's existing paper repository or can be retrieved on-the-fly if open-access or partner links exist for these works. These matching papers are parsed, and pulled into our agent state for usage with other tools. Note that the search services will have access to a larger corpus of works than is available to us via our repository and accessible link traversal, in these cases the system will simply skip these papers and they are not used. A stub of the paper search tool is implemented in \texttt{paperqa} with directions for users to implement their own retrieval since it will be limited to their own access to full-text papers. 

Ablations and configurations for workflows like WikiCrow are exposed in \texttt{paperqa} as nested configuration objects. All experiments performed in this work correspond to included configuration objects. Here we highlight the configuration variable descriptions corresponding to the salient features tested in this work, though all variable names are available in the included files. 

\begin{figure}[h!]
    \centering 
    \refstepcounter{suppfigure}
    \includegraphics[width=0.9\textwidth]{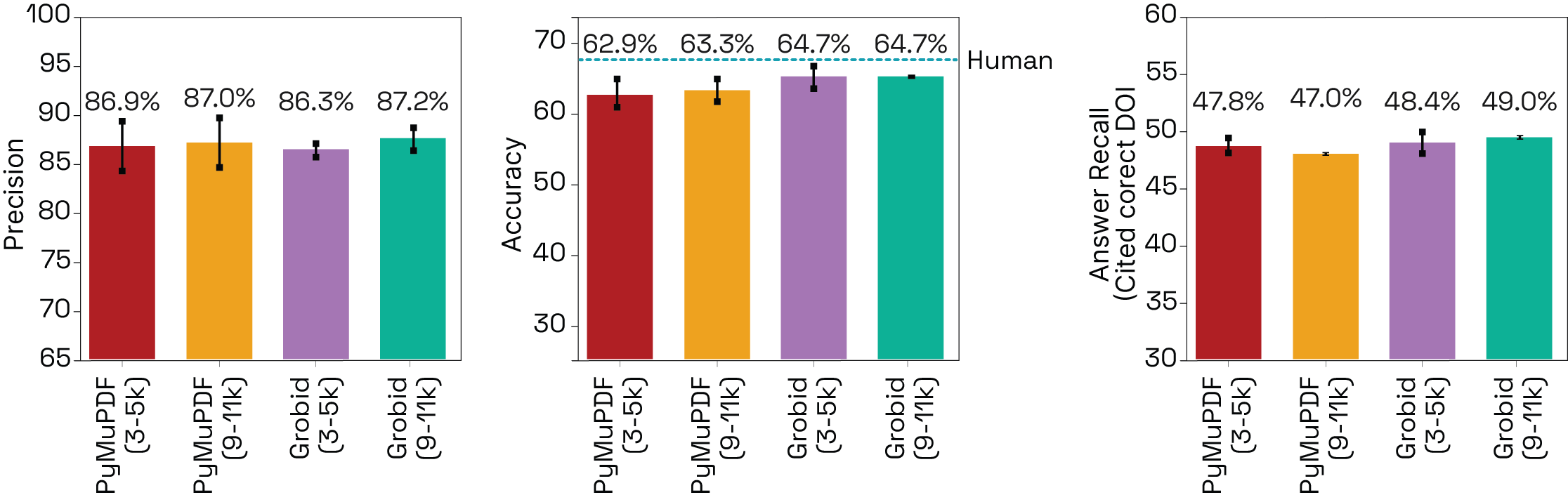}
    \caption{Precision, accuracy and recall for different chunk size and parsing algorithm choices on LitQA. Note that the gap between recall and precision is because the model can sometimes find an alternative source to rule out distractors, provide the same information, or make the model confident enough to guess.} 
    \label{fig:chunksize_ablations} 
\end{figure}

\begin{itemize}
\item \texttt{query}: The main query task asked of the PaperQA agent, i.e. a LitQA question or a directive to write an article. 
\item \texttt{llm}: The LLM used in the generate answer tool can be a valid Anthropic, OpenAI, or Gemini model identifier. This parameter was varied for the model experiments in \autoref{fig:litqa_performance}.
\item \texttt{agent\_llm}: The LLM used for the agent orchestration, in this work, it was always fixed to \texttt{gpt-4-turbo-2024-04-09}.
\item \texttt{summary\_llm}: The LLM used for the RCS step in the gather evidence tool, must be a valid Anthropic, OpenAI, or Gemini model identifier. This parameter was varied for the model ablations in \autoref{fig:litqa_performance}.
\item \texttt{prompts}: A \texttt{PromptCollection} object from PaperQA \cite{lala2023paperqa}, which allows for specification of prompts in each tool, as well as features like turning off RCS (via \texttt{prompts.skip\_summarization}). The \textit{No RCS Model} ablation used this input as well as the WikiCrow prompts.
\item \texttt{max\_sources}: The number of top ranked sources to be included in the generate answer tool, in \textbf{\autoref{fig:paperqa_summary}A}, the `filter top summaries'' cutoff. This parameter was 5 in the top-performing \textit{Answer cutoff @ 5}, but 15 for all other experiments.
\item \texttt{consider\_sources}: The top-k cutoff, i.e. the number of chunks that will be used in the RCS step. This parameter was set to 30 by default in LitQA experiments, save for the \textit{Top-k Rank @ X} experiment where it was set to X. Additionally, for our WikiCrow prompts this parameter was set to 25. 
\item \texttt{agent\_tools}: An ordered list of tool names that will be used by the agent, including \texttt{gather\_evidence}, \texttt{paper\_search}, \texttt{generate\_answer}, and \texttt{citations\_traversal}. This always included all four tools except for the \textit{No Cit. Trav.} and \textit{No Agent} runs where \texttt{citations\_traversal} was excluded.
\item \texttt{docs\_index\_mmr\_lambda}: A pre-gather evidence MMR lambda parameter which can be used to pre-filter similar papers by name before gathering evidence. This was set to 0.9 for our WikiCrow run to promote diversity of sources, but 1.0 for LitQA experiments. 
\item \texttt{parsing\_configuration.ordered\_parser\_preferences}: A list of the parsing algorithm to use, either \texttt{paperqa\_default} (PyMuPDF) or \texttt{grobid}. \texttt{paperqa\_default} was the default for each ablation, and \texttt{grobid} was used for WikiCrow generation. This parameter was also varied in the experiments shown in \textbf{Figure \ref{fig:chunksize_ablations}}. 
\item \texttt{parsing\_configuration.chunksize}: the chunk size (in characters) to be used when chunking parsed documents. This parameter was varied in the experiments shown in \textbf{Figure \ref{fig:chunksize_ablations}}.
\item \texttt{parsing\_configuration.overlap}: the overlap (in characters) that will be common between sequential chunks. This was fixed at 750 for this work. 
\item \texttt{parsing\_configuration.chunking\_algorithm}: the algorithm used to chunk documents, \texttt{simple\_overlap} simply uses a sliding window with overlap, and \texttt{sections} uses semantic parsing by section (i.e. one chunk per section where possible), if sections need to be broken into multiple chunks the system will automatically handle this. \texttt{sections} is only supported via \texttt{parsing\_configuration.ordered\_parser\_preferences}=\texttt{grobid}. This parameter was varied in the experiments shown in {Figure \ref{fig:chunksize_ablations}}, and in our WikiCrow generation.
\item \texttt{temperature}: temperature used for the LLM in the generate answer tool.  This was set to 0 for all runs in this work.
\item \texttt{summary\_temperature}: temperature used for the LLM in the gather evidence tool's RCS step, this was set to 0 for all runs in this work.
\end{itemize}

\subsubsection{Tool implementations} \label{sec:cit-trav}

PaperQA2's agentic tools were implemented as in PaperQA\cite{lala2023paperqa}. Our agent was prompted with the following message to guide tool usage: 

\begin{quote}
Answer question: \{question\}. Search for papers, gather evidence, collect papers cited in evidence then re-gather evidence, and answer. Gathering evidence will do nothing if you have not done a new search or collected new papers. If you do not have enough evidence to generate a good answer, you can:
\\
- Search for more papers (preferred)\\
- Collect papers cited by previous evidence (preferred)\\
- Gather more evidence using a different phrase\\

If you search for more papers or collect new papers cited by previous evidence, remember to gather evidence again. Once you have five or more pieces of evidence from multiple sources, or you have tried a few times, call \{gen\_answer\_tool\_name\} tool. The \{gen\_answer\_tool\_name\} tool output is visible to the user, so you do not need to restate the answer and can simply terminate if the answer looks sufficient. The current status of evidence/papers/cost is \{status\}
\end{quote}

Where variables like \{status\} are included to represent the current state to the agent. Tools were implemented with the following prompts and settings. \\

\textbf{Paper Search Tool} \\

The paper search tool uses an initial keyword search, generated by the agent in the context of the user query. The agent is prompted as follows: 

\begin{quote}
A search query in this format: [query], [start year]-[end year]. You may include years as the last word in the query, e.g. 'machine learning 2020' or 'machine learning 2010-2020'. The current year is \{get\_year()\}. The query portion can be a specific phrase, complete sentence, or general keywords, e.g. 'machine learning for immunology'.
\end{quote}

Our initial search relies on services like Semantic Scholar\cite{semantic-scholar}, where candidate lists (default of 12) of relevant papers are generated then parsed. When parsed, the papers are first turned into text using either Grobid or PyMuPDF, then split into \texttt{chunksize} character sized pieces. If the \texttt{sections} parsing is used, then section chunks are split on header metadata provided by Grobid. An embedding vector is generated for each chunk using a hybrid implementation which concatenates a dense and sparse, keyword based embedding model. For the experiments included in this study, OpenAI's \texttt{text-embedding-large-3} was used. It was concatenated with a normalized 256 dimension vector which used modulus-encoding to extract a hot-encoded keyword from the tokenization integers provided by OpenAI's tiktoken \cite{tiktoken} library. These text chunks are put into a document context which is accessible by the agent for further manipulation with tools. The PaperQA entrypoint for these functions can be found on \href{https://github.com/Future-House/paper-qa}{github}. \\ \\

\textbf{Gather Evidence Tool} \\

As detailed in \href{https://github.com/Future-House/paper-qa}{github} for PaperQA, the Gather Evidence tool begins with a top-k vector ranking step, using the embedding vectors created in the Paper Search tool. The user query is embedded with the same algorithm, and cosine similarity is used to match all document chunks in the agent's context with the user query. The top-k chunks are then selected for the RCS step. 

The reranking and contextual summarization step most differentiates PaperQA's implementation relative to other RAG technologies. The tool's prior step, an top-k vector retrieval ranking, is a widely implemented \cite{ma2023queryrewritingretrievalaugmentedlarge, gao2023retrieval} approach to identify relevant documents, however, the RCS second step, is unique to PaperQA (to the authors' knowledge). While performance improvements with both deep reranking (or LLM) models and map-reduced summarizations \cite{sarthi2024raptor, zhuang2023opensourcellm, gao2023chatrec, yu2024rankrag} are well documented, combining the reranking operation with a contextual summary provides novel benefits. 

The step is implemented by mapping an LLM completion across each top-k chunk (system prompt):
\begin{quote}
Provide a summary of the relevant information that could help answer the question based on the excerpt. The excerpt may be irrelevant.  Do not directly answer the question - only summarize relevant information. Respond with the following JSON format:
\{\{ "summary": "...",  \\
"relevance\_score": "..."
\}\}
where "summary" is relevant information from text - \{summary\_length\} words and "relevance\_score" is the relevance of "summary" to answer the question (integer out of 10)
\end{quote}

Where each chunk is injected as follows:
\begin{quote}
    Excerpt from {citation}
    ----
    \{text\}
    ----
    Query: \{question\}
\end{quote}

After completion, each JSON object is parsed and the passages are re-ranked according to the new relevance scores. When running with WikiCrow, gene names are also prompted to be extracted as additional JOSN keys, these are kept and injected in the final answering context. Advantages of the RCS step are as follows: 1. Token usage efficiency is vastly improved, a contextual summary will be 200-400 tokens compared with our standard document's chunk size of 2,250 tokens. This allows for a significantly more accessible document corpus for injection into PaperQA's answering context window. Furthermore, we see no decrease in summarization efficacy, using LitQA performance as a proxy, across document chunk sizes from 750-3,000 tokens. 2. As a new feature in this work, the LLM can be prompted to provide its summary in a structured JSON or XML format to simplify its downstream data extraction. In addition to a relevance score used for reranking, this structure can include metadata (such as a gene name) which will be retained through the PaperQA workflow. This is used to reduce hallucination and confusion in the final answer context. Since the RCS step is performed in an embarrassingly parallel fashion, it's highly efficient, and its utility can be applied to an arbitrarily deep ranking, up to the rate or cost limits of the LLM API. Our studies on the efficacy of the RCS depth led us to use a much deeper RCS depth, and to utilize the best performing model for the RCS operation. This differs from the intuition in prior work \cite{lala2023paperqa}, which utilized a cheaper model during the RCS step. \\

\textbf{Generate Answer Tool} \\

This tool answers questions by taking a subset of the top ranked sources (from the RCS ranking), and injects them into a final context for answering. The default in this study was to inject 15 contextual summaries, but we saw maximal accuracy with 5 at the cost of precision. LLMs were prompted to answer as follows:

\begin{quote}
Answer the question below with the context.

Context:
\{context\}
----
Question: \{question\}

Write an answer based on the context. If the context provides insufficient information and the question cannot be directly answered, reply "I cannot answer." For each part of your answer, indicate which sources most support it via citation keys at the end of sentences, like (Example2012Example pages 3-4). Only cite from the context and only use the valid keys. Write in the style of a Wikipedia article, with concise sentences and coherent paragraphs. The context comes from a variety of sources and is only a summary, so there may inaccuracies or ambiguities. If quotes are present and relevant, use them in the answer. This answer will go directly onto Wikipedia, so do not add any extraneous information.

Answer (\{answer\_length\}):
\end{quote}

Where contexts are injected by the generate answer code before output is returned to the agent. \\

\textbf{Citation Traversal Tool} \\

Atop the PaperQA\cite{lala2023paperqa} tools, we created an additional tool to traverse one degree of citations, both forward in time (``future citers'') and backwards in time (``past references''). This tool enables a fine-grained search around paper(s) containing relevant information. The traversal originates from any paper containing a highly-scored contextual summary (RCS score 0-10), and our minimum score threshold was eight (inclusive). The papers corresponding to highly-scored summaries are referred to as $D_\text{prev}$ in \textbf{Algorithm \ref{alg:cit-trav}}. See \textbf{\autoref{tab:cit-trav-starting-dist}} for the frequencies of various $|D_\text{prev}|$ when this tool was selected.

\begin{table}[htb]
\caption{Various statistics on citation traversal.}
\begin{subtable}{\textwidth}
\centering
\begin{tabular}{| c || c c c c c c c c c c |}
\hline
$|D_\text{prev}|$ & 1 & 2 & 3 & 4 & 5 & 6 & 7 & 8 & 9 & 10 \\
\hline
Count & 2147 & 941 & 530 & 386 & 307 & 216 & 154 & 67 & 29 & 12 \\
\hline
Frequency (\%) & 44.8 & 19.6 & 11.1 & 8.1 & 6.4 & 4.5 & 3.2 & 1.4 & 0.6 & 0.3 \\
\hline
\end{tabular}
\caption{Distribution of traversal starting paper count $|D_\text{prev}|$.}
\label{tab:cit-trav-starting-dist}
\end{subtable}

\begin{subtable}{\textwidth}
\centering
\begin{tabular}{| c || c c c c c c c |}
\hline
$|D_\text{prev}|$ & 1 & 2 & 3 & 4 & 5 & 6 & 7+ \\
\hline
Frequency of 1 Overlap (\%) & \textbf{100.0} & \textbf{91.3} & \textbf{91.7} & 90.8 & 90.8 & 88.7 & 87.5 \\
\hline
Frequency of 2 Overlaps (\%) & \cellcolor{gray!25} & \textbf{8.7} & \textbf{7.4} & \textbf{7.5} & \textbf{7.2} & \textbf{8.2} & 8.5 \\
\hline
Frequency of 3 Overlaps (\%) & \cellcolor{gray!25} & \cellcolor{gray!25} & \textbf{0.9} & \textbf{1.4} & \textbf{1.4} & \textbf{2.0} & \textbf{2.4} \\
\hline
Frequency of 4 Overlaps (\%) & \cellcolor{gray!25} & \cellcolor{gray!25} & \cellcolor{gray!25} & \textbf{0.3} & \textbf{0.5} & \textbf{0.7} & \textbf{0.9} \\
\hline
Frequency of 5 Overlaps (\%) & \cellcolor{gray!25} & \cellcolor{gray!25} & \cellcolor{gray!25} & \cellcolor{gray!25} & \textbf{0.1} & \textbf{0.3} & \textbf{0.4} \\
\hline
Frequency of 6 Overlaps (\%) & \cellcolor{gray!25} & \cellcolor{gray!25} & \cellcolor{gray!25} & \cellcolor{gray!25} & \cellcolor{gray!25} & \textbf{0.1} & \textbf{0.2} \\
\hline
Frequency of 7+ Overlaps (\%) & \cellcolor{gray!25} & \cellcolor{gray!25} & \cellcolor{gray!25} & \cellcolor{gray!25} & \cellcolor{gray!25} & \cellcolor{gray!25} & \textbf{0.1} \\
\hline
\end{tabular}
\caption{Table showing the frequencies of citation overlap $o$ seen in LitQA, illustrating the percentage of traversed citations at stake when filtering with an overlap threshold $\theta_o$ . We chose to specify $\theta_o = \lceil \alpha \times |D_\text{prev}| \rceil$, where $\alpha$ is known as the overlap fraction and was defaulted to $\frac{1}{3}$. The bolded values show what overlaps would have been preserved using an $\alpha = \frac{1}{3}$.}
\label{tab:cit-trav-overlap-dist}
\end{subtable}
\end{table}

To first acquire citations, Semantic Scholar \cite{semantic-scholar} and Crossref \cite{crossref} APIs are called for past references and Semantic Scholar APIs are called for future citers. To collect all citations for a given paper, we make one API call per provider per direction, totalling four API calls/paper. All three providers only provide partial paper details, meaning a large fraction of the time a title or DOI is not present in the response metadata. To merge citations across providers, a best-effort de-duplication is performed using casefolded title and lowercased DOI. In \textbf{Algorithm \ref{alg:cit-trav}}, this logic takes place inside the \texttt{GetCitations} procedure.

Once citations have been acquired, bins of overlap $\mathcal{B}$ are computed. For example, traversing past references for the following six DOIs:
\texttt{10.1016/j.mcn.2006.08.007}, \texttt{10.1002/cpsc.17}, \texttt{10.1002/(sici)1098- 1136(200004)30:2<105::aid-glia1>3.0.co;2-h}, \texttt{10.1089/scd.2015.0244}, \texttt{10.1002/glia.22882}, and \texttt{10.1042/an20120041},
leads to one DOI cited by four papers, five DOIs cited by three papers, 29 DOIs cited by two papers, and 428 DOIs cited by just one paper.

To filter bins of overlap, a hyperparameter ``overlap fraction'' $\alpha$ was introduced to compute a threshold overlap $\theta_o$ as a function of the number of source papers ($|D_\text{prev}|$). For example, with an $\alpha = \frac{2}{5}$ and traversing from six source DOIs, all citations not appearing in at least three source DOIs were discarded. The default overlap fraction used in data collection was $\frac{1}{3}$. See \textbf{\autoref{tab:cit-trav-overlap-dist}} for a full distribution of overlaps seen during LitQA runs. Furthermore, a twelve paper limit $\ell$ was posed on the traversal, which meant in the above example only keeping six of the bin of 29 DOIs cited by two papers. To filter within a bin, we fall back on the count of future citers. This winnowing logic is detailed across \textbf{Algorithm \ref{alg:cit-trav}}'s \texttt{FilterOverlap} and \texttt{TraverseCitations} procedures. Lastly, we traverse both future citers and past references, feathering together the resultant DOIs before finding them. 


\begin{algorithm}
\setstretch{1.15}
\caption{Traverse Citations}
\label{alg:cit-trav}
\begin{algorithmic}[1]

\Require{Set of summaries $S$, score threshold $\theta_\text{score}$, overlap fraction $\alpha$, look future flag $\mathds{1}_\text{fut}$, limit $\ell$}
\Ensure{Set of traversed papers $D_\text{out}$, where papers are future citers if $\mathds{1}_\text{fut}$ else past references}
\Procedure{TraverseCitations}{$S$, $\theta_\text{score}$, $\alpha$, $\mathds{1}_\text{fut}$, $\ell$}
\State $D_\text{prev} \leftarrow \{s_d \mid s \in S \wedge s_\text{score} \geq \theta_\text{score} \}$
\Comment{Traverse from highly-scored summaries' corresponding papers}
\State $\mathcal{D} \leftarrow \Call{GetCitations}{D_\text{prev}, \mathds{1}_\text{fut}}$
\Comment{$\mathcal{D}$ is a set of sets of papers such that $|\mathcal{D}| = |D_\text{prev}|$}
\State $\theta_o \leftarrow \lceil \alpha \times |\mathcal{D} | \rceil$
\Comment{Overlap threshold $\theta_o$ scales with $|S|$}
\State \Return \Call{FilterOverlap}{$\mathcal{D}$, $D_\text{prev}$, $\theta_\text{o}$, $\ell$}
\EndProcedure

\Require{Set of sets of candidate papers $\mathcal{D}$, set of previous papers $D_\text{prev}$, (inclusive) overlap threshold $\theta_o$, limit $\ell$}
\Ensure{Set of filtered papers $D_\text{out}$}
\Procedure{FilterOverlap}{$\mathcal{D}$, $D_\text{prev}$, $\theta_\text{o}$, $\ell$}
    \State $\mathcal{B} = \left[ \left(o, \{d \mid \left( \sum_{D \in \mathcal{D}} \mathds{1} (d \in D) \right) = o \}\right) \: \textbf{for} \: o \in \left[ |\mathcal{D}|, \dots, 1 \right] \right]$
    \Comment{Bin papers according to decreasing overlap}
    \State $D_\text{out} \leftarrow \{\}$
    \For{$o, D \in \mathcal{B}$}
    \Comment{Highest overlapping citations come first}
        \If{ $o < \theta_o \vee | D_\text{out} | \geq \ell$ }
            \textbf{break}
        \EndIf
        \State $D \leftarrow \{d \mid d \in D \wedge d \notin D_\text{prev} \}$
        \Comment{Filter out already present papers}
        \If {$\ell - |D_\text{out}| < |D|$}
        \Comment{If the entire bin won't fit within limit $\ell$}
            \State $D \leftarrow \{d \mid i, d \in $ \Call{$\text{sort}_{\downarrow \text{citers}}$}{$D$} $\wedge \: i \leq ( \ell - | D_\text{out} | ) \}$
            \Comment{Keep subset with the most future citers}
        \EndIf
        \State $D_\text{out} \leftarrow D_\text{out} \cup D$
    \EndFor
    \State \Return $D_\text{out}$
\EndProcedure

\end{algorithmic}
\end{algorithm}

\subsection{LitQA}

\subsubsection{Question Construction and Human Evaluation} \label{sec:question_evaluators}

LitQA questions were generated manually by a combination of the authors as well as contracted human experts (see \cite{laurent2024lab}). All human annotators were compensated, informed that their evaluations were being used in research of human-level performance, and consented to the use of their annotations and participation. Question authors were instructed to identify recent papers (published within the last ~36 months), and develop a multiple-choice question that requires context within the main text of the paper to answer and \textit{is not answerable by the abstract or title alone.} They were further advised that the question should require some amount of reasoning within the paper context, and not be a direct quote or statement from the paper. Distractors were instructed to be reasonable within the context, using either other information in the paper (e.g. other genes being discussed) or based on inherent or other knowledge. We also periodically tested question drafts with ChatGPT 3.5 or 4 (with logging disabled) to ensure questions were not easily answerable by models already, or to help design effective distractors by asking to provide plausible answers. Question drafts were also often searched against Google Scholar to ensure that it was not trivial to find the exact statement necessary to answer the question. This was an effective aid in revising question wording.

For contracted question authors or anyone new to drafting, the first 10 or so questions produced were carefully reviewed by one or more of the authors for quality, after which they were either asked to rework them, they were added to corpus, or they were removed from consideration. This feedback (referred to as \verb|calibration|) was usually enough to ensure quality question generation going forward from contractors, though questions were reviewed by authors on an ongoing basis prior to merging into the main corpus. Contractors were compensated via multiple structures during different phases of generation as we iterated on effective strategies. Initially, they were paid on an hourly basis at \$50 per hour. At later phases, they were paid on a completion basis such that they were paid per accepted question, which equated to variable hourly equivalents always \$50 or more per hour. Using this methodology, LitQA2 was built up from LitQA (47 questions) in two stages of releases, first 100 questions (147), then an additional 101 questions, adding to the original subset to make 248 total questions. 

Human evaluators were assigned questions from the LitQAv2 corpus in rounds of "quizzes" composed of 20-40 questions each. They were allocated up to a week to complete the quiz, but were given no other time constraints. They were also allowed to use tools such as internet search or journal collection search provided via their institutions. They were asked to explicitly refrain from using AI-based tools such as ChatGPT or Claude, though we did not have any method of enforcement of this request. In order to encourage completion as well as high performance, we attempted to design an incentive scheme that separately promoted both: In brief, we paid a base dollar amount (from \$3-6) to each question for completion. We then added a bonus to each question that was based on overall performance: 
\begin{itemize}
    \item{80\% or more overall score = bonus equal to base question amount for each correct question.}
    \item{60\% - 80\% overall score = bonus equal to half the base question amount for each correct question.}
    \item{Less than 60\% overall score = bonus equal to \$1 for each correct question.}
\end{itemize}

We additionally paid a completion bonus for the entire quiz in some instances, amounting to \$150. On average, evaluators were paid between \$50 to \$100 per hour based on self-estimation of time spent and total compensation. Quizzes were sometimes combined with similar evaluation quizzes for other evaluation benchmark categories we are developing.

For the human LitQA2 performance reported in this work, 2 rounds of quizzes (20 questions each) were given to 9 evaluators. In total, evaluators answered provided 266 unique answers across 248 unique questions.

A third round of quizzes was given with another set of 160 questions given to evaluators, with questions which overlapped from both the initial set of 147 questions and the remaining 101 questions. However, this quiz round had several outlier scores ( >90\%) in quizzes on the same subsets of questions which had been given previously to a much lower average (66.3\%). Upon investigating, we found that a portion of the initial 147 LitQA2 questions that were available on GitHub had been parsed by google-indexed data aggregators, causing it to trivially return results with a Google search from evaluators. Interestingly, this did not impact PaperQA2 performance, because PaperQA2 does not use Google websearch. Thus, we chose to exclude evaluator answers to the initial 147 LitQA2 questions in the third set of quizzes as it was no longer a valid human comparison.

\subsubsection{PaperQA measurement}
When measuring LitQA2 performance, \texttt{paperqa} function calls were performed for each LitQA question. LitQA answer order was randomized for each call, and, for all configurations shown, 3 full runs of all LitQA questions were performed. 3 runs were necessary to control for the inherent noise in LLM inference, even at 0 temperature (where all tests were performed).  LitQA2 was automatically evaluated using an evaluation LLM call (GPT-4-0613), which extracted the letter answer from PaperQA2's output. It was prompted to extract as follows: 

\begin{quote}
Extract the single letter answer from the following question and answer
\{prompt\_output(`QA')\}

Single Letter Answer:
\end{quote}

The extraction was then parsed and graded against the LitQA2 benchmarks correct ideal answer. All questions were graded as ``Unsure'' if the ``Insufficient information to answer this question'' option was selected, except for the LitQA2 questions where the ideal answer is ``null'', then the ``Insufficient information'' was graded as ``Correct''. 

LitQA2 was developed in two stages of 100 then 101 questions, adding to the original subset of LitQA (47) questions to make 248 questions. This gave us an opportunity to evaluate our system before and after adding the set of 101 previously unseen questions to evaluate overfitting. We can see that our system performed similarly in terms of accuracy between question sets, implying we were not overfitting to the original LitQA data. 

\begin{table}[h]
\centering
\begin{tabular}{|c|c|c|c|c|}
\hline
Ablation & Accuracy (Old 147) & Accuracy (New 101) & Precision (Old 147) & Precision (New 101) \\ \hline
Claude-3-Opus & 59.0\% $\pm $2.4\% & 54.5\% $\pm $6.1\% & 93.9\% $\pm $3.7\% & 82.6\% $\pm $6.0\% \\ \hline
Gemini-1.5-Pro & 55.8\% $\pm $5.7\% & 59.4\% $\pm $2.3\% & 93.2\% $\pm $1.8\% & 83.0\% $\pm $4.6\% \\ \hline
GPT-4-Turbo &  64.4\% $\pm $1.8\% & 63.7\% $\pm $7.4\% & 91.3\% $\pm $2.1\% & 81.8\% $\pm $7.3\% \\ \hline
\end{tabular}
\caption{Table of model choices using both stages of LitQA2 development, all errors shown are CI intervals.}
\label{tab:litqa_before_after}
\end{table}

When comparing PaperQA2 to other RAG technologies in \textbf{Figure  \ref{fig:litqa_performance}A}, the author's manually entered each LitQA2 question (in the exact format given to PaperQA2) into the online interfaces for both Perplexity\cite{perplexity} and Elicit\cite{elicit}. Perplexity was run using the "Pro" product, and GPT-4o, across all 248 LitQA2 questions. We recorded either the letter answer, or manually interpreted the response to assign the system's letter output. The performance metrics were calculated in an identical way to PaperQA. Elicit was not able to be run for all 248 questions, the numbers reported only represent the first tranche of 147 LitQA2 questions. On Elicit's final 101 LitQA2 questions, the authors noted a significant algorithm change. Answers resulted in > 90 \% "unsure" answers, thus we are only reporting on the subset of LitQA2 where multiple choice answers could be extracted. We also compared PaperQA2 to PaperQA's prior published configuration\cite{lala2023paperqa}, which used a smaller RCS model (GPT 3.5 Turbo) and ranking depth, had no citation traversal tool, and did not have access to Grobid parsings. PaperQA's LitQA2 accuracy was 36.7\% and its precision was 76.5\%. Thus, PaperQA2 showed a large improvement in both metrics, particularly in accuracy.

\subsection{WikiCrow} \label{sec:si-wikicrow}

WikiCrow statements were generated using a linked set of 5 different queries, 4 to PaperQA2, and 1 to a frontier LLM model, GPT-4-Turbo. As detailed in \textbf{Figure \ref{fig:WikiCrow_results}A}, each PaperQA2 query created a different article section. The PaperQA2 prompts and settings used for these queries can be found in our \texttt{paperqa} \href{https://github.com/Future-House/paper-qa}{repository}. The primary PaperQA2 parameter differentiation between WikiCrow and LitQA2 runs was the usage of Grobid for document parsing, as well as the modification of the RCS step to extract a gene name that is the focus of each paper chunk being summarized. They were designed to reduce hallucinations and accurately read data from tables.

We compared WikiCrow and Wikipedia statements starting from a sample of 300 genes which were sampled from from a complete list of all protein encoding human genes \cite{enwiki:proteinlist} (n=19,255) after being filtered for only those having valid (non-stub) Wikipedia articles (n=3,639). Among the 300 articles, Wikipedia statements were sampled, first from the pool of 300 genes, then among each article's \texttt{<p>} HTML sections with more than 25 characters. 500 total Wikipedia statements were sampled, which resulted in 240 unique genes. Matching WikiCrow articles were generated for each of the 240 genes for comparison. WikiCrow statements were broken at either a reference list (notated with parenthesis and PaperQA2 document names) or at a paragraph break, most closely matching Wikipedia's HTML structure. Each of the 500 Wikipedia statements were randomly matched with a WikiCrow statement from the same gene article to ensure that evaluators were given a similar gene-distribution of statements between WikiCrow and Wikipedia articles. 

References were extracted from the Wikipedia statements using the \texttt{<sup>} link, and matched to the correct DOI or hyperlink in the article references. References were extracted from the WikiCrow statements by parsing the parenthetical insertions (via regular expressions) and matching the DOI to the document name given by the model. In this way, each statement was associated with a list of DOI links to the sources listed. Efforts were then made to structure each statement similarly, removing any HTML from the Wikipedia statements, and removing all inserted references into the WikiCrow statements. All references were replaced with random numbers between 1 and 30, but kept consistent if listed several times in the same statement. References were then uniformly reformatted as \texttt{[x]}, to hide their origin.

The matching 500 Wikipedia and 500 WikiCrow statements were shuffled for each of the 4 evaluators. Each was given a list of 200 to grade while blinded to the statement origins. Graders were sent the questions with the following instructions: 

\begin{quote}
\textbf{Task overview}\\ \\
You’ll be reviewing statements written about human genes. These statements are intended to provide information taken from the literature, and should thus provide an accurate citation such that the stated information matches information contained in the cited literature. You’ll be evaluating each statement according to its accuracy as well as being cited appropriately. Please read the guidelines below carefully to ensure accurate work.\\ \\
\textbf{Scoring} \\

You’ll soon receive a link to a private Google Sheet with 200 statements, along with some associated information for each one, including links to any cited literature. You’ll be grading the sentences in each statement according to the below metrics. Some statements may contain multiple sentences, each of which should be evaluated such that any sentence not satisfying our scoring criteria counts against the entire statement. \\
\begin{itemize}
\item{Is it cited? In other words, does the statement contain one or more citations to published literature?}
\item{Is the information correct, as cited? In other words, is the information stated in the sentence correct according to the literature that it cites?\\}
\end{itemize}

There will be a column for each of the above scores. Put TRUE in the appropriate column if true according to the metric, and FALSE if not. There will also be a column for “notes”. Please use this column to include anything you think is important for your grading, or other flags we might want to know about. Below are some more detailed guidelines for scoring:

You are only evaluating if the statements are correct as cited -- not if the statement is globally correct or supported by other sources. Some common scenarios you may encounter:

\begin{itemize}

\item{\textbf{Simple statements:} Statement has a single citation and the content is supported by the citation. Is Cited = TRUE, Is Info Correct As Cited = TRUE}

\item{\textbf{Simple unsupported statements:} A statement is true, highly-specific (see "broad context" below, but not explicitly stated in the source, mark as FALSE for “correct as cited”, but mark "is cited" as TRUE.}

\item{\textbf{Unrelated, meta statements:} Statement is a meta-statement or unrelated to biology. For example, the statement may be the start of a list like: "Gene XYZ has the following properties:", or an attribution like "This data has been provided by Y organization." These statements should be judged as "Not Applicable/NA". They do not need to be cited.}

\item{\textbf{Broad context:} Statement has a citation and the content is a broader context than the citation, but not explicitly explained in the citation. For example, an explanation of what a secondary structure is or what a protein sequence is. The broad content should be judged to be "undergraduate biology student common knowledge". If it is, then the info can be marked correct as cited, if it's more advanced than undergraduate bio knowledge (or outright incorrect), then the info is incorrect as cited.}

\item{\textbf{Underspecified citation:} The statement may be factually correct, but the citation is under-specified. An example might be that the citation is a link to a database, where a user needs to query the database and synthesize their own data to support the statement. If it requires synthesis by the user, than the statement should be judged as NOT correct as cited. If it's a database link to the a page about the gene, and the fact can be found there, then it can be judges as correct as cited. }

\item{\textbf{Inaccessible citation:} The statement may be factually correct, but the citation is to an inaccessible source, like a closed source database or a textbook. These should be judged as "Not Applicable/NA". Though an attempt should be made to acquire the source to validate the information.}

\item{\textbf{Multi-part citations:} The statement has several sentences and citations. Each sentence should be correct as cited given all of the previous criteria (including the "Broad contexts" category).}

\item{Occasionally, it may not be clear from context which sentence applies to a source, there may be three sentences in a row with two sources trailing the sentences. In this case if the sentences are supported by either source, then it's correct as cited. If a single sentence is un-supported by either source, and they don't meet our broad context criteria, then mark as incorrect as cited.}

\item{If there are multiple citations, where one is inaccessible, but the accessible citations can support the statements by themselves, then mark as correct as cited. If the statement can NOT be validated by the accessible sources, then mark as "Not Applicable/NA".}
\end{itemize}
\end{quote}

Evaluators were incentivized to encourage both quality of review and quantity of statements reviewed. They were offered \$10 per statement graded as a base payment, and were additionally offered a \$10 per question bonus payment if they reviewed more than 50 questions by a specific deadline approximately five days after starting.

Using this criteria, 4 expert researchers were contracted as evaluators to grade 375 evaluations. They were instructed to only complete their statements in the order provided, and any statements that were not graded in a contiguous block from the initial statement were excluded from analysis (to avoid biasing towards simple statements). Among our 375 evaluations, 40 statements overlapped between 2 or more evaluators. Among those 40, 77.5\% (31) had agreement, and 22.5\% (9) had a disagreement in evaluation, showing good overall alignment in scoring. 

We had previously reported\cite{coxWikiCrow} lower ``cited and unsupported'' percentages for both Wikipedia and WikiCrow, but the prior study differed in key ways: 1. this work limits the statement samples to be from the same subset of gene articles, eliminating selection bias for longer, higher quality articles; 2. this work uses statements built from sentences which may cite one or more sources as is typically present in articles, the prior work focused on only single sentences with one source which greatly limited the sample; and 3. this work used a much more robust evaluation pipeline, with blinded external evaluators.

Due to the complexity and time required to evaluate WikiCrow accuracy, we relied on heuristics to evaluate our summarization-centric features, adding metadata (gene name) extraction to our RCS step and Grobid-based parsings. These features were meant to mitigate gene name confusion between the RCS step and the generate answer tool, and misreading tabular data, which anecdotally were the two most frequent issues found by the authors while developing WikiCrow. Among the reasoning issues reported by the evaluators, only 2 / 171 statements had issues with gene name confusion, and there were no reported table extraction errors. Utilizing Grobid structured parsings also reduced PaperQA2's token usage. Across a sample of 5,363 papers, PyMuPDF's parsings resulted in an average of 16,040 tokens per paper, while Grobid's parsings resulted in an average 8,903 tokens. The savings can be attribution to a reduction in excess whitespace as well as the removal of reference sections, which we did not include in our Grobid parsings.

\subsection{Contradiction Detection Methods}  \label{sec:cc-methods}
ContraCrow detects contradictions in literature through two steps. Unless provided, the first step in ContraCrow is to extract claims from a given paper \textbf{\autoref{fig:contradiction_detection}A}. First, the provided paper is split into chunks. To maintain context for extracting claims, each chunk is only split within its respective section, up to 5000 characters (no overlap), and the section and paper titles are included. Each resulting chunk is then fed into a claim-extraction LLM to extract claims candidates. The resulting extracted claims are each evaluated by a filtering LLM that assesses the claims based on quality and generalizability, assigning a score out of 10. Only claims scoring 8 or higher are considered. 

Once claims have been generated or are provided, each claim is fed into PaperQA2 with a special \textit{contradiction detection prompt}, instructing the system to search for contradictions and respond with an appropriate format. The system considers each claim independently and identifies relevant papers across literature in order to identify possible contradictions to the claim. Finally, as instructed by the \textit{contradiction detection prompt}, ContraCrow outputs its reasoning and a choice from an 11-point Likert scale (see \autoref{fig:contradiction_detection}D). We map the natural language score to integers (0-10) in our calculations, which allows us to tune the decision threshold, as seen in \ref{fig:contradiction_detection}B. For all experiments, this step uses Claude 3.5 Sonnet \cite{claude2024} as the LLM model, a chunk size of 7000 characters, a temperature of 0, and a simple overlap of 250. Unless otherwise specified, we use a decision boundary of $8$ on this step. 

\subsubsection{ContraDetect} \label{sec:cc-contradetect}
The ContraDetect benchmark was generated from the LitQA2 contradiction dataset, removing any questions with ``null'' responses as to not overlap with the no-evidence data. The resulting questions were randomly split into two groups. The questions and corresponding ideal answers from the first group were individually given to \texttt{gpt-4-turbo-2024-09}\cite{achiam2023gpt} with instructions to rephrase them into factual statements. The second group was similarly turned into incorrect statements (contradictions), given question and the first corresponding distractor answers. Each question corresponds to exactly 1 statement. These claims then proceeded into ContraCrow's second step, as described above.

We separately designed 42 no-evidence statements by hand. The authors carefully designed questions in their respective fields of expertise and did extensive literature review to ensure that no literature has reported on the claim. There is no overlap between the no-evidence statements and the LitQA2 questions. The resulting answers from ContraCrow were also evaluated by the authors to ensure that no evidence had been found. Any claims with found evidences were removed from this dataset. 

\subsubsection{Human Evaluation} \label{sec:cc-eval}
We randomly selected 100 papers from our local database of biology papers. 7 of these papers failed to parse, leaving our dataset at 93 papers. These papers have no overlap with the papers corresponding to LitQA2 questions. \texttt{gpt-4-turbo-2024-04-09}\cite{achiam2023gpt} was used for both LLM models in the claim-generation step. The resulting claims were then fed into the contradiction-detection step as described above, in batches of 1,000 claims.

For the human ``contradiction validation'' task, 50 claims scored at 8 and 50 claims at least 9 by ContraCrow were randomly chosen (distribution: $8:50$, $9:48$, $10:2$). The resulting 100 contradictions were split over 5 expert evaluators evenly, with no overlap. The evaluators were provided with the claim, the original claim information (source paper reference and the chunk used to generate the claim), some background information (acronyms, definitions, etc.), ContraCrow's reasoning, and all chunks cited in the reasoning. The evaluators were tasked with giving two separate binary responses indicating their agreement with both the model's reasoning and conclusion, based on the provided cited chunks (up to 10). They were instructed to use additional provided resources sparingly, only for clarification, and to \textit{only} consider the evidence chunks and ContraCrow's reasoning when making a final decision. Only the label for ``agreement with the model's conclusion'' was used for analysis. 

For the human ``contradiction detection'' task, 10 claims scored 0-4, 10 claims scored 8, and 10 claims scored at least 9 by ContraCrow were randomly chosen (distribution: $0:1$, $1:6$, $2:2$, $3:1$, $8:10$, $9:10$). The resulting 30 claims were split over 5 evaluators. Each claim was evaluated by at least 2 evaluators. The evaluators were provided with the claim, the original claim information (source paper reference and the chunk used to generate the claim), background information (acronyms, definitions, etc.), and all literature chunks considered by ContraCrow (up to 15). The evaluators were tasked with assigning a Likert scale value and a binary (``yes'' or ``no'') label to determine whether each claim contained a contradiction to the provided sources. For this task, the evaluators did not have access to ContraCrow's reasoning. Similar to the first task, they were instructed to use any provided materials for clarification, but to only use the provided literature when making a final decision. 

In the both the ``contradiction validation'' task and the binary ``contradiction detection'' task, ContraCrow scores of $>=8$ were considered the positive case in order to compare to humans' binary outputs. For the Likert score tasks, raw scores were used.

Evaluators were compensated for their work on grading claims to encourage thorough review. For the ``contradiction validation'' task, they were paid \$12 per claim completed. For ``contradiction validation'', they were paid \$20 per claim completed. They were additionally paid a \$200 `completion bonus' for completing grading on all claims by a deadline approximately one week after starting.

Evaluators were sent the questions with the following instructions:  \\
\begin{quote}
\textbf{Contradiction Detection Instructions} \\
\textbf{Task overview}\\ 
For this task, we are interested in identifying and rating contradicting research findings, or claims. We’ve assembled a set of claims from research papers that are potentially contradicted by other findings, either prior or later. Your task will be to both assess the claim itself, as well as review portions of text taken from research papers that may be related to and may or may not present contradicting information to the presented claim.
We will not have an explicit calibration phase for this task, nor do we provide examples. More detailed instructions are provided below to help guide you. You may use any sources available to you in order to understand concepts, acronyms, etc. that you are unfamiliar with, but you should perform the evaluation itself by-hand and using only the claim and context provided.
We also explicitly ask that you do not use ChatGPT or other AI tools to come to your conclusions, though they may be used if helpful to define terms or clarify concepts. Please do not put claims or chunks into ChatGPT and similar tools.

\textbf{Instructions}\\
This task will consist of two separate activities, Claim Evaluation and Contradiction Detection (type a and b). For each claim, you will be given the following information:
\begin{itemize}
    \item Claim: a single sentence making some claim or finding.
    \item Chunk: the section of text from the source paper that was used to generate the claim.
    \item Section: the section of the paper that the chunk lives in, if available
    \item Title: the title of the paper that the chunk lives in, if available
    \item DOI: the DOI URL for the paper that the chunk lives in, if available
\end{itemize}

\textbf{Section 1: Claim Evaluation}\\
Your first task is to a determination of the claim being a valid claim. Some examples of good \& bad claims are shown below:
\begin{itemize}
    \item Good claims are typically clear, testable, and generalizable
    \begin{itemize}
        \item ``Cells expressing ARG1 are more likely to undergo apoptosis.''
    \end{itemize}
    \item Bad claims will come in different forms, like not having enough context, i.e. it does not make sense outside the context of the paper.
    \begin{itemize}
        \item ``Cells in cluster 1 had higher expression of ARG1.''
        \item ``Cells expressing this gene are more likely to undergo apoptosis.''
        \item ``Method A performed better than Method B in our study.''
        \item Claims provided may also just not really be defined as a `claim'. These might describe methods, descriptions, or obvious common knowledge
            \begin{itemize}
                \item ``We used method A to test the effects of ARG1 on cell death.''
                \item ``ARG1 is a protein.''
            \end{itemize}
        \item Claims may also not be supported by the provided chunk (Mixed up gene names, off-topic discussion, etc.)
    \end{itemize}
\end{itemize}
To evaluate each claim, select all that apply from the following options:
\begin{itemize}
    \item good claim
    \item not enough context
    \item not generalizable
    \item not a claim
    \item not supported
\end{itemize}

Some important considerations:
\begin{itemize}
    \item Multiple claims may come from the same chunk \& claims may be redundant in nature
    \item Consider each claim independently
    \item You may use the all provided information in this section to help you on this step (including the paper title, as it may provide context)
    \item Avoid:
    \begin{itemize}
        \item Evaluating the claim for truth value
        \item Evaluating the wording/syntax of the claim, unless it changes the meaning
        \item Deep-diving into the paper outside of the claim
        \item Spending more than a few minutes grading each claim, on average
    \end{itemize}
\end{itemize}

\textbf{Section 2: Contradiction Detection}\\
This part will contain additional information for each claim, consisting of up to 15 ``evidence chunks'' which are sections of research paper text that may serve as evidence either supporting, contradicting or neutral to the claim to various degrees.
Each chunk will itself contain (when available) the source paper's citation. This information can be used for reference and clarity (for defining acronyms, clarifying things, etc) but should not be considered when determining contradictions.
Your job for this part of the task is to determine whether (and to what extent) the claim contradicts or agrees with the provided evidence chunks. You will provide a selection for both of the below categories (note there are two separate selections!)
\begin{itemize}
    \item Contradiction grade (choose one of the following):
        \begin{itemize}
            \item Explicit Agreement
            \item Strong Agreement
            \item Agreement
            \item Possibly an Agreement
            \item Lack of Evidence
            \item Possibly a Contradiction
            \item Nuanced Contradiction
            \item Contradiction
            \item Strong Contradiction
            \item Explicit Contradiction
        \end{itemize}
    \item Contradiction determination:
        \begin{itemize}
            \item YES - Claim contradicts the provided context
            \item NO - Claim does not contradict the provided context
        \end{itemize}
\end{itemize}
Some important considerations:
\begin{itemize}
    \item A claim may have multiple chunks from the same paper.
    \item Consider claims individually. If the claim is a bad claim, use your best judgement about whether or not and how to evaluate the contradiction. We ask that you attempt to evaluate it if possible, and if you feel you cannot proceed, please explain why then move on to the next claim.
\item Focus on the claim + context provided: Don't worry about other contradictions or agreements in the context, only focus on the specific claim. Additional information (publication information, background info) is provided for clarification where needed, but you should rely solely on the provided context and claim for labeling.
\item It is possible that the provided statement makes more than 1 claim. Use your best judgement in this case, considering the context provided.
\item Avoid:
    \begin{itemize}
        \item Evaluating the claim or evidences for truth value or quality (only evaluate in relation to each other).
        \item Deep-diving into any text outside of the provided context (including the papers from which the context was derived).
    \end{itemize}
\end{itemize}
\end{quote}

\begin{quote}
\textbf{Contradiction Validation Instructions }\\
\textbf{Task overview}\\
This task is very similar, but this time you will be given an AI model's reasoning on the evidence chunks, and only the evidence chunks the model considered (which should cut down the review time considerably). You will do the usual annotation, and additionally say whether you agree with the LLM's reasoning. Section 1 is unchanged, Section 2 is different.

We also explicitly ask that you do not use ChatGPT or other AI tools to come to your conclusions, though they may be used if helpful to define terms or clarify concepts. Please do not put claims or chunks into ChatGPT and similar tools.
\\
\textbf{Instructions}\\
This task will consist of two separate activities, Claim Evaluation and Contradiction Detection (type a and b). For each claim, you will be given the following information:
\begin{itemize}
    \item Claim: a single sentence making some claim or finding.
    \item Chunk: the section of text from the source paper that was used to generate the claim.
    \item Section: the section of the paper that the chunk lives in, if available
    \item Title: the title of the paper that the chunk lives in, if available
    \item DOI: the DOI URL for the paper that the chunk lives in, if available
\end{itemize}

\textbf{Section 1: Claim Evaluation}\\
\textit{\*\*Section 1 is the same as in the ``contradiction detection'' task.\*\*}

\textbf{Section 2: Contradiction Detection - Model Evaluation}\\
This part will contain additional information for each claim. Each claim will have a model response, which is some reasoning to answer the question `Are there any contradictions to this claim in literature?' The model response will reference up to 10 “evidence chunks” which are sections of research paper text.
These referenced chunks will also be provided for you and can be linked via the citation key in the model response (for example, the model response might reference \texttt{Song2022DEPTH2:pages 16-17}. The chunk corresponding to this reference will also be found under \texttt{Song2022DEPTH2:pages 16-17}.
Each chunk will itself contain (when available) the source paper's citation. This information can be used for reference and clarity (for defining acronyms, clarifying things, etc) but should not be considered when evaluating contradictions or model response.
Your job for this task is to determine whether you agree with the model response and to what extent the claim contradicts or agrees with the evidence chunks provided.
You will provide a selection for both of the below categories (note there are two separate selections!)
\begin{itemize}
    \item Contradiction grade (choose one of the following):
        \begin{itemize}
            \item Explicit Agreement
            \item Strong Agreement
            \item Agreement
            \item Possibly an Agreement
            \item Lack of Evidence
            \item Possibly a Contradiction
            \item Nuanced Contradiction
            \item Contradiction
            \item Strong Contradiction
            \item Explicit Contradiction
        \end{itemize}
    \item Contradiction determination:
        \begin{itemize}
            \item YES - Claim contradicts the provided context
            \item NO - Claim does not contradict the provided context
        \end{itemize}
    \item Optional Explanation: optionally, you may provide a written explanation/justification of your scoring.
\end{itemize}

Some important considerations:
\begin{itemize}
    \item A claim may have multiple chunks from the same paper.
    \item Consider claims individually.
    \item Because the model response is provided, we ask that you evaluate every provided claim.
    \item Additional information (publication information, background info) is provided for clarification where needed.
    \item It is possible that the provided statement makes more than one claim. Use your best judgment in this case, considering the model response provided.
    \item Avoid:
    \begin{itemize}
        \item Evaluating the claim or evidences for truth value or quality (only evaluate in relation to each other).
        \item Evaluating the model response truth value, quality, syntax, or style, except where directly related to the claim.
        \item Deep-diving into any text outside of the provided context (including the papers from which the context was derived).
    \end{itemize}
\end{itemize}
\end{quote}

\bibliographystyle{unsrt}  
\bibliography{references}

\end{document}